\documentclass{article}

\usepackage[ruled,vlined, linesnumbered]{algorithm2e}

\usepackage[usenames,dvipsnames]{color}
\usepackage{bbm}
\usepackage{enumitem}

\usepackage{algpseudocode}
\let\oldnl\nl
\newcommand{\nonl}{\renewcommand{\nl}{\let\nl\oldnl}}

\usepackage{bm}
\usepackage{amsmath,amssymb,amsthm}

\usepackage{microtype}
\usepackage{graphicx}
\usepackage{subfigure}
\usepackage{booktabs} 
\newcommand{\tra}[1]{\renewcommand{\arraystretch}{#1}}

\usepackage{multirow}

\usepackage{pifont}

\usepackage{mathtools}

\usepackage{wrapfig}

\DeclarePairedDelimiter\abs{\lvert}{\rvert}%

\usepackage{hyperref}

\usepackage[final]{neurips_2021}

\newcommand{\autoattack}{\scode{AA}\xspace}
\newcommand{\tool}{\scode{A}$^3$\xspace} 

\newcommand{\dataset}{\ensuremath{\mathcal{D}}}

\newcommand{\norm}[1]{\ensuremath{\lVert #1 \rVert}}

\newcommand{\scode}[1]{{\fontfamily{lmtt}\selectfont
#1}}
\newcommand{\sbcode}[1]{{\color{blue}\ensuremath{\mathtt{#1}}}}

\newcommand{\apgdce}{\scode{APGD}$_{\texttt{CE}}$}
\newcommand{\apgddlr}{\scode{APGD}$_{\texttt{DLR}}$}

\newcommand{\captiona}{{\em (a)}}
\newcommand{\captionb}{{\em (b)}}
\newcommand{\captionc}{{\em (c)}}

\newcommand{\captionn}[1]{{\em (#1)}}

\def\Figref#1{Figure~\ref{#1}}
\def\Tabref#1{Table~\ref{#1}}

\def\Secref#1{Section~\ref{#1}}

\def\eqref#1{eq.~\ref{#1}}
\def\Eqref#1{Eq.~\ref{#1}}

\def\Algref#1{Algorithm~\ref{#1}}

\def\1{\bm{1}}

\def\vx{{\bm{x}}}

\DeclareMathAlphabet{\mathsfit}{\encodingdefault}{\sfdefault}{m}{sl}
\SetMathAlphabet{\mathsfit}{bold}{\encodingdefault}{\sfdefault}{bx}{n}

\def\gA{{\mathcal{A}}}

\def\gH{{\mathcal{H}}}

\def\gS{{\mathcal{S}}}

\def\sL{{\mathbb{L}}}

\def\sR{{\mathbb{R}}}
\def\sS{{\mathbb{S}}}
\def\sT{{\mathbb{T}}}

\def\sX{{\mathbb{X}}}

\def\sZ{{\mathbb{Z}}}

\newcommand{\E}{\mathbb{E}}

\newcommand{\R}{\mathbb{R}}

\DeclareMathOperator*{\argmax}{arg\,max}

\usepackage{colortbl}

\usepackage{tikz}
\usetikzlibrary{positioning}
\usetikzlibrary{calc}
\usetikzlibrary{decorations.pathreplacing}
\usetikzlibrary{arrows.meta}
\usetikzlibrary{plotmarks}
\usetikzlibrary{snakes,automata,backgrounds,petri}
\usepackage{tikz-qtree}
\usetikzlibrary{matrix}
\usetikzlibrary{arrows,shapes,patterns}
\tikzset{
    >=stealth',
    boxx/.style={
           rectangle,
           draw=black,
           text width=6.5em,
           minimum height=2em,
           text centered},
}

\begin{document}



\title{Automated Discovery of Adaptive Attacks on Adversarial Defenses}

%

\author{
 Chengyuan Yao \\
 Department of Computer Science \\
 ETH Z\"{u}rich, Switzerland \\ 
 \texttt{chengyuan.yao@inf.ethz.ch} \\
 \And
 Pavol Bielik \\
 LatticeFlow\\
 Switzerland \\
 \texttt{pavol@latticeflow.ai} \\
  \And
 Petar Tsankov \\
 LatticeFlow\\
 Switzerland \\
 \texttt{petar@latticeflow.ai} \\
  \And
 Martin Vechev \\
 Department of Computer Science \\ 
 ETH Z\"{u}rich, Switzerland \\
 \texttt{martin.vechev@inf.ethz.ch} \\
}


\maketitle

\begin{abstract}
Reliable evaluation of adversarial defenses is a challenging task, currently limited to an expert who manually crafts attacks that exploit the defense’s inner workings or approaches based on an ensemble of fixed attacks, none of which may be effective for the specific defense at hand. Our key observation is that adaptive attacks are composed of reusable building blocks that can be formalized in a search space and used to automatically discover attacks for unknown defenses. We evaluated our approach on 24 adversarial defenses and show that it outperforms \scode{AutoAttack}~\citep{croce2020reliable}, the current state-of-the-art tool for reliable evaluation of adversarial defenses: our tool discovered significantly stronger attacks by producing 3.0\%-50.8\% additional adversarial examples for 10 models, while obtaining attacks with slightly stronger or similar strength for the remaining models.


\end{abstract}

\section{Introduction}
%

The issue of adversarial attacks~\citep{szegedy2013intriguing, goodfellow2014explaining}, i.e., crafting small~input perturbations that lead to mispredictions, is an important problem with a large body of recent~work.
%
Unfortunately, reliable evaluation of proposed defenses is an elusive and challenging task: many defenses seem to initially be effective, only to be circumvented later by new attacks designed specifically with that defense in mind~\citep{adversarial_not_easy_to_detect, athalye2018obfuscated, tramer2020adaptive}.

To address this challenge, two recent works approach the problem from different perspectives. \citet{tramer2020adaptive} outlines an approach for manually crafting adaptive attacks that exploit the weak points of each defense. Here, a domain expert starts with an existing attack, such as \scode{PGD}~\citep{madry2017towards} (denoted as $\bullet$ in \Figref{overview}), and adapts it based on knowledge of the defense's inner workings.
Common modifications include: \captionn{i} tuning attack parameters (e.g., number of steps), \captionn{ii} replacing network components to simplify the attack (e.g., removing randomization or non-differentiable components), and \captionn{iii} replacing the loss function optimized by the attack. This approach was demonstrated to be effective in breaking all of the considered $13$ defenses. However, a downside is that it requires substantial manual effort and is limited by the domain knowledge of the expert -- for instance, each of the $13$ defenses came with an adaptive attack which was insufficient, in retrospect.

At the same time, \citet{croce2020reliable} proposed to assess adversarial robustness using an ensemble of four attacks illustrated in \Figref{overview}~\captionb{} -- \apgdce{} with cross-entropy loss~\citep{croce2020reliable}, \apgddlr{} with difference in logit ratio loss, \scode{FAB}~\citep{croce2019minimally}, and \scode{SQR}~\citep{andriushchenko2019square}. While these do not require manual effort and have been shown to improve the robustness estimate for many defenses, the approach is inherently limited by the fact that the attacks are fixed apriori without any knowledge of the given defense at hand. This is visualized in \Figref{overview}~\captionb{} where even though the attacks are designed to be diverse, they cover only a small part of the entire~space.

\DeclareRobustCommand\searchbox{\tikz{\draw[fill=green!30!white] (6.69,0) rectangle ++(0.2,0.2);}}

\begin{figure*}[t]
\centering
\begin{tikzpicture}[auto,trim left=0cm, scale=1.2]
\tikzset{attack/.style= {font=\scriptsize, text width=2cm, align=center}}
\tikzset{blank/.style= {font=\scriptsize}}

\begin{scope}[xshift=3.9cm]

\draw[draw=black] (0,0) rectangle ++(3.8,2);

\node[attack] at (1,0.6) {$\bullet$ \\ \apgdce{}};
\node[attack] at (0.6,1.1) {$\bullet$ \\ \apgddlr{}};
\node[attack] at (1.7,1.0) {$\bullet$ \\ \scode{FAB}};
\node[attack] at (1.2,1.5) {$\bullet$ \\ \scode{SQR}};





\node[font=\footnotesize, text width=4.5cm, align=center] at (1.9, 2.3) {\captionb{} Ensemble of fixed attacks\\\cite{croce2020reliable}};

\node[font=\footnotesize] at (2.0, -0.18) {$\bullet$~fixed attack};

\end{scope}

\begin{scope}

\node[font=\footnotesize, text width=4.9cm, align=center] at (1.8, 2.3) {\captiona{} Handcrafted adaptive attacks\\\cite{tramer2020adaptive}};

\draw[->] (2.35,-0.2) to [out=30,in=150] (2.65,-0.2);
\node[font=\footnotesize] at (3.3, -0.2) {manual step};
\node[font=\footnotesize] at (1.1, -0.2) {$\triangleright$~best adaptive attack};

\draw[draw=black] (0,0) rectangle ++(3.8,2);

\node[attack] at (1.1,0.6) {$\bullet$ \\ \scode{PGD}};
\node[attack] at (0.5,1.6) {  ensemble \\ diversity \\ {\small $\triangleright$}};

\node[attack] (bab) at (0.3,0.3) {$\bullet$ \\ \scode{B\&B}};

\draw[->] (1.05,0.8) to [out=120,in=-30] node [xshift=0.8cm, yshift=-0.4cm, text width=2cm, midway,left, font=\scriptsize, align=center] {optimize \\ params} (0.55,1.3);
\draw[->] (0.25,0.5) to [out=120,in=-120] (0.45,1.3);

\node[attack] (nes-a) at (1.2,1.5) {$\bullet$ \\ \scode{NES}};
\node[blank] (nes-b) at (1.9,1.4) {$\circ$};
\node[attack] (kwta) at (2.7,1.5) {{\small $\triangleright$} \\ \scode{k-WTA}};

\draw[->] (1.3,1.6) to [out=-0,in=180] node [yshift=0cm, midway,above, font=\scriptsize] {$\uparrow n$} (1.8,1.4);
\draw[->] (1.95,1.45) to [out=30,in=180] node [yshift=0.1cm, midway,above, font=\scriptsize] {+modified loss} (2.6,1.6);

\node[blank] (empir-a) at (1.8,0.7) {$\circ$};
\node[attack] (empir) at (2.5,0.5) {{\small $\triangleright$} \\ \scode{EMPIR}};

\draw[->] (1.2,0.76) to [out=30,in=150] node [yshift=0.1cm, midway,above, font=\scriptsize] {+BPDA} (1.7,0.7);
\draw[->] (1.9,0.7) to [out=30,in=150] node [yshift=-0.1cm, midway,above, font=\scriptsize] {+weighted loss} (2.4,0.64);

\node[blank] at (1.5,0.5) {$\circ$};
\node[blank]  at (1.8,0.4) {$\circ$};
\node[attack] at (2.1,0.15) {{\small $\triangleright$}};

\draw[->] (1.2,0.72) to [out=30,in=150] (1.45,0.54);
\draw[->] (1.55,0.48) to [out=30,in=150] (1.75,0.42);
\draw[->] (1.85,0.38) to [out=30,in=120] (2.05,0.22);

\end{scope}

\begin{scope}[xshift=7.8cm]

\node[font=\footnotesize, text width=4.3cm, align=center] at (1.9, 2.3) {\captionc{} Adaptive attack search\\ (Our Work)};

\draw[->, dashed] (2.2,-0.2) -- (2.5,-0.2);
\node[font=\footnotesize] at (3.1, -0.2) { search step};

\node[font=\footnotesize] at (-1.4, -0.2) {\searchbox{} search space};

\draw[draw=black] (0,0) rectangle ++(3.8,2);

\draw[fill=green!30!white] plot[smooth, tension=.7] coordinates {
(0.4, 0.3) (0.2, 0.7) (0.2, 1.1) (0.3, 1.7) (1.2, 1.8) (2.3, 1.65) (3.4, 1.8) (3.7, 1.5) (3.6, 1.0) (2.9, 1.0) (2.6, 0.8) (2.9, 0.3) (1.5, 0.35) (0.7, 0.2) (0.4, 0.3) 
};

\node[attack] at (1,0.6) {$\bullet$ \\ \scode{PGD}};
\node[attack] at (0.6,1.1) {$\bullet$ \\ \scode{APGD}};
\node[attack] at (1.7,1.0) {$\bullet$ \\ \scode{FAB}};
\node[attack] at (3.3,1.4) {$\bullet$ \\ \scode{SQR}};
\node[attack] at (1.2,1.5) {$\bullet$ \\ \scode{NES}};
\node[attack] at (2.3,1.1) {$\bullet$ \\ \scode{DeepFool}};
\node[attack] at (0.5,0.7) {$\bullet$ \\ \scode{C\&W}};
\node[attack] at (1.2,0.95) {$\bullet$ \\ \scode{FGSM}};

\node[attack] at (2.1,0.15) {{\small $\triangleright$}};
\node[attack] (empir) at (2.5,0.5) {{\small $\triangleright$} \\ \scode{EMPIR}};
\node[attack] (kwta) at (2.7,1.5) {{\small $\triangleright$} \\ \scode{k-WTA}};
\node[attack] at (0.5,1.6) {  $ $ \\ $ $ \\ {\small $\triangleright$}};

\draw[->, dashed] (1.1, 0.7) -- (1.4, 0.7) -- (1.7, 0.6) -- (2.0, 0.8) -- (2.4, 0.6);
\draw[->, dashed] (0.7, 1.25) -- (0.8, 1.35) -- (0.6, 1.45);
\draw[->, dashed] (1.3, 1.6) -- (1.6, 1.65) -- (1.9, 1.4) -- (2.2, 1.5) -- (2.6, 1.6);

\draw[->, dashed] (1.1, 0.7) -- (1.4, 0.5) -- (1.8, 0.5) -- (1.95, 0.35);


\end{scope}

\end{tikzpicture}

\vspace{-1em}
\caption{High-level illustration and comparison of recent works and ours. \emph{Adaptive attacks} \captiona{} rely on a human expert to manually adapt an existing attack to exploit the weak points of each defense. \scode{AutoAttack}~\captionb{} evaluates defenses using an ensemble of diverse attacks. Our work \captionc{} defines a search space of adaptive attacks (denoted as \searchbox{}) and performs search steps automatically. 
}
\label{overview}
\end{figure*}
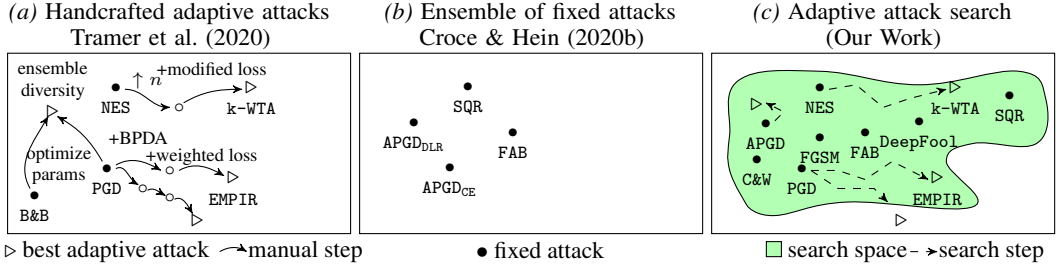

\paragraph{This work: towards automated discovery of adaptive attacks}
We present a new method that helps automating the process of crafting adaptive attacks, combining the best of both prior approaches; the ability to evaluate defenses automatically while producing attacks tuned for the given defense.
Our work is based on the key observation that we can identify common techniques used to build existing adaptive attacks and extract them as reusable building blocks in a common framework. Then, given a new model with an unseen defense, we can discover an effective attack by searching over suitable combinations of these building blocks.
To identify reusable techniques, we analyze existing adaptive attacks and organize their components into three groups:

\begin{itemize}[leftmargin=*]
\item  \emph{Attack algorithm and parameters}: a library of diverse attack techniques (e.g., \scode{APGD}, \scode{FAB}, \scode{C\&W}~\citep{CandW}, \scode{NES}~\citep{wierstra2008natural}), together with backbone specific and generic parameters (e.g., input randomization, number of steps, if/how to use \scode{EOT}~\citep{athalye2018obfuscated}).

\item \emph{Network transformations}: producing an easier to attack surrogate model using techniques including variants of \scode{BPDA}~\citep{athalye2018obfuscated} to break gradient obfuscation, and layer removal~\citep{tramer2020adaptive} to eliminate obfuscation layers such as redundant softmax operator.

\item \emph{Loss functions}: that specify different ways of defining the attack's loss function.
\end{itemize}

These components collectively formalize an attack search space induced by their different combinations. We also present an algorithm that effectively navigates the search space so to discover an attack. In this way, domain experts are left with the creative task of designing new attacks and growing the framework by adding missing attack components, while the tool is responsible for automating many of the tedious and time-consuming trial-and-error steps that domain experts perform manually today. That is, we can automate some part of the process of finding adaptive attacks, but not necessarily the full process. This is natural as finding truly new attacks is a highly creative process that is currently out of reach for fully automated techniques.

We implemented our approach in a tool called \scode{Adaptive} \scode{AutoAttack} (\tool{}) and evaluated it on $24$ diverse adversarial defenses. Our results demonstrate that \tool{} discovers adaptive attacks that outperform \scode{AutoAttack}~\citep{croce2020reliable}, the current state-of-the-art tool for reliable evaluation of adversarial defenses: \tool{} finds attacks that are significantly stronger, producing 3.0\%-50.8\% additional adversarial examples for 10 models, while obtaining attacks with stronger or simialr strength for the remaining models. Our tool \tool{} and  scripts for reproducing the experiments are available online at: 
\begin{center}
\url{https://github.com/eth-sri/adaptive-auto-attack}
\end{center}


\section{Automated Discovery of Adaptive Attacks}

We use $\dataset = \{(x_i, y_i)\}_{i=1}^N$ to denote a training dataset where $x \in \sX$ is a~natural input (e.g., an image) and $y$ is the corresponding label. An adversarial example is a perturbed input~$x'$, such that: \captionn{i} it satisfies an attack criterion~$c$, e.g., a $K$-class classification model $f\colon \sX \rightarrow \sR^{K}$ predicts a wrong label, and \captionn{ii} the distance $d(x', x)$ between the adversarial input~$x'$ and the natural input~$x$ is below a~threshold $\epsilon$ under a distance metric~$d$ (e.g., an $L_p$ norm). Formally, this can be written as:

\vspace{-0.5em}
\begin{figure}[h]
\hspace{7em}
\begin{tikzpicture}
\node at (-1,0) {
\begin{minipage}{0.4\textwidth}
\begin{equation*}
\displaystyle
d(x', x) \leq \epsilon \quad \text{such that} \quad c(f, x', x)
\end{equation*}
\end{minipage}
};

\draw [decorate,decoration={brace,amplitude=4pt},xshift=0pt,yshift=2pt]
(-3.3,0.2) -- (-2.1,0.2) node [black,midway,yshift=0.4cm, text width=4cm, align=center] {\footnotesize capability};

\draw [decorate,decoration={brace,amplitude=4pt},xshift=0pt,yshift=2pt]
(0.1,0.2) -- (1.5,0.2) node [black,midway,yshift=0.4cm, text width=4cm, align=center] {\footnotesize goal (criterion)};

\node[text width=2cm, align=center] at (-5.2,0.0) {\footnotesize
\textsc{Adversarial\\Attack}
};

\end{tikzpicture}
\vspace{-0.5em}
\end{figure}

For example, instantiating this with the $L_{\infty}$ norm and misclassification criterion corresponds to:

\vspace{-0.5em}
\begin{figure}[h]
\hspace{7em}
\begin{tikzpicture}

\node at (-1,0) {
\begin{minipage}{0.4\textwidth}
\begin{equation*}
\displaystyle
\norm{x' - x}_{\infty} \leq \epsilon \;\;\; \text{s.t.} \;\;\; \hat{f}(x') \neq \hat{f}(x)
\end{equation*}
\end{minipage}
};

\node[text width=2.8cm, align=center] at (-5.2,0) {\footnotesize
\textsc{Misclassification\\$L_{\infty}$ Attack}
};
\end{tikzpicture}
\vspace{-1.0em}
\end{figure}



where $\hat{f}$ returns the prediction $\argmax_{k=1:K} f_k(\cdot)$ of the model $f$.

%
%
%
%



\paragraph{Problem Statement} Given a model $f$ equipped with an unknown set of defenses and a dataset $\dataset = \{(x_i, y_i)\}_{i=1}^N$, our goal is to find an adaptive adversarial attack $a \in \gA$ that is best at generating adversarial samples $x'$ according to the attack criterion $c$ and the attack capability $d(x', x) \leq \epsilon$:
\begin{equation}\label{problem_statement}
\max_{a \in \gA,~d(x', x) \leq \epsilon} \E_{(x, y) \sim \dataset}\quad c(f, x', x) \qquad \qquad \text{where} \quad x' = a(x, f)
\end{equation}
Here, $\gA$ denotes the search space of all possible attacks, where the goal of each attack $a\colon \sX \times (\sX \rightarrow \sR^K) \rightarrow \sX$ is to generate an adversarial sample $x' = a(x,f)$ for a given input~$x$ and model~$f$.
For example, solving this optimization problem with respect to the $L_{\infty}$ misclassification criterion corresponds to optimizing the number of adversarial examples misclassified by the model.

In our work, we consider an \emph{implementation-knowledge adversary}, who has full access to the model's implementation at inference time (e.g., the model's computational graph).
We chose this threat model as it matches our problem setting -- given an unseen model implementation, we want to automatically find an adaptive attack that exploits its weak points but without the need of a domain expert.
We note that this threat model is weaker than a \emph{perfect-knowledge adversary}~\citep{biggio2013evasion}, which assumes a domain expert that also has knowledge about the training dataset 
and algorithm, as this information is difficult, or even not possible, to recover from the model's implementation only.

\paragraph{Key Challenges}
To solve the optimization problem from \Eqref{problem_statement}, we address two key challenges:

\begin{itemize}[leftmargin=*]
\item \textit{Defining a suitable attacks search space $\gA$} such that it is expressible enough to cover a range of existing adaptive attacks.
\item \textit{Searching over the space $\gA$ efficiently} such that a strong attack is found within a reasonable time.
\end{itemize}

Next, we formalize the attack space in \Secref{search_space} and then describe our search algorithm in \Secref{search}.

\section{Adaptive Attacks Search Space}\label{search_space}

We define the adaptive attack search space by analyzing existing adaptive attacks and identifying common techniques used to break adversarial defenses. Formally, the adaptive attack search space is given by $\gA\colon \sS \times \sT$, where $\sS$ consists of sequences of backbone attacks along with their loss functions, selected from a space of loss functions $\sL$, and $\sT$ consists of network transformations.
Semantically, given an input $x$ and a model $f$, the goal of adaptive attack $(s,t)\in \sS \times \sT$ is to return an adversarial example $x'$ by computing $s(x, t(f)) = x'$. That is, it first transforms the model $f$ by applying the transformation~$t(f) = f'$, and then executes the attack $s$ on the surrogate model $f'$.
Note that~the surrogate model is used only to compute the candidate adversarial example, not to evaluate it. That is, we generate an adversarial example~$x'$ for~$f'$, and then check whether it is also adversarial for~$f$. Since $x'$ may be adversarial for~$f'$, but not for~$f$, the adaptive attack must maximize the transferability of the generated candidate adversarial~samples.

\paragraph{Attack Algorithm \& Parameters ($\sS$)}
The attack search space consists of a sequence of adversarial attacks. We formalize the search space with the grammar:

\begin{center}
{
\fontfamily{lmtt}\selectfont

\begin{tabular}{@{}llll@{}}
\multicolumn{2}{l}{(Attack Search Space)} \\
 $\qquad \sS$ ::= & $\sS$; $\sS$ $\mid$
\sbcode{randomize} $\sS$ $\mid$ \sbcode{EOT} $\sS$, n $\mid$ \sbcode{repeat} $\sS$, n $\mid$ \\ & \sbcode{try} $\sS$ \sbcode{for} n $\mid$ Attack \sbcode{with} params \sbcode{with} loss\;$\in \sL$ \\
\end{tabular}
}
\end{center}

\begin{itemize}
\item $\sS;~\sS$: composes two attacks, which are executed independently and return the first adversarial sample in the defined order. That is, given input~$x$, the attack $s_1;s_2$ returns $s_1(x)$ if $s_1(x)$ is an adversarial example, and otherwise it returns $s_2(x)$.

\item \sbcode{randomize}~$\sS$: enables the attack's randomized components, which correspond to~random seed and/or selecting a starting point within $~d(x', x) \leq \epsilon$, uniformly at random.

\item \sbcode{EOT}~$\sS$\scode{, n}: uses expectation over transformation, a technique designed to compute gradients for models with randomized components~\citep{athalye2018obfuscated}.

\item \sbcode{repeat}~$\sS$\scode{, n}: repeats the attack $n$ times (useful only if randomization is enabled).

\item \sbcode{try}~$\sS$~\sbcode{for}~\scode{n}: executes the attack with a time budget of \scode{n} seconds.

\item \scode{Attack}~\sbcode{with}~\scode{params} \sbcode{with} \scode{loss}\;$\in \sL$: is a backbone attack \scode{Attack} executed with parameters \scode{params} and loss function \scode{loss}. In our evaluation, we use \scode{FGSM}~\citep{goodfellow2014explaining}, \scode{PGD}, \scode{DeepFool}~\citep{moosavi2016deepfool}, \scode{C\&W}, \scode{NES}, \scode{APGD}, \scode{FAB} and \scode{SQR}. 
We provide full list of the attack parameters, including their ranges and priors in Appendix \ref{sec_search_space}.
\end{itemize}

Note, that we include variety of backbone attacks, including those that were already superseeded by stronger attacks.
This is done for two key reasons.
First, weaker attacks can be surprisingly effective in some cases and avoid the detector because of their weakness (see defense \scode{C24} in our evaluation).
Second, we are not using any prior when designing the space search. In particular, whenever a new attack is designed it can simply be added to the search space. Then, the goal of the search algorithm is to be powerful enough to perform the search efficiently.
In other words, the aim is to avoid making any assumptions of what is useful or not and let the search algorithm learn this instead.

%

\paragraph{Network Transformations ($\sT$)}

A common approach that aims to improve the robustness of neural networks against adversarial attacks is to incorporate an explicit defense mechanism in the neural architecture. These defenses often obfuscate gradients to render iterative-optimization methods ineffective~\citep{athalye2018obfuscated}. However, these defenses can be successfully circumvented by \captionn{i} choosing a suitable attack algorithm, such as score and decision-based attacks (included in $\sS$), or \captionn{ii} by changing the neural architecture (defined next).

At a high level, the network transformation search space $\sT$ takes as input a model $f$ and transforms it to another model~$f'$, which is easier to attack. To achieve this, the network $f$ can be expressed as a directed acyclic graph, including both the forward and backward computations, where each vertex denotes an operator (e.g., convolution, residual blocks, etc.), and edges correspond to data dependencies.
In our work, we include two types of network transformations:

\textit{Layer Removal}, which removes an operator from the graph. Each operator can be removed if its input and output dimensions are the same, regardless of its functionality.

\textit{Backward
Pass Differentiable Approximation} (\scode{BPDA})~\citep{athalye2018obfuscated}, which replaces the backward version of an operator with a differentiable approximation of the function. In our search space we include three different function approximations: \captionn{i} an identity function, \captionn{ii} a convolution layer with kernel size 1, and \captionn{iii} a two-layer convolutional layer with ReLU activation in between. The weights in the latter two cases are learned through approximating the forward function.


\paragraph{Loss Function ($\sL$)}
Selecting the right objective function to optimize is an important design decision for creating strong adaptive attacks.
Indeed, the recent work of \citet{tramer2020adaptive} uses 9 different objective functions to break 13 defenses, showing the importance of this step.
We formalize the space of possible loss functions using the following grammar:

{
\fontfamily{lmtt}\selectfont

\begin{tabular}{@{}rll@{}}
\multicolumn{3}{l}{(Loss Function Search Space)} \\[0.3em]
$\sL$ ::= & 
\sbcode{targeted} Loss, n \sbcode{with} Z $\mid$ \sbcode{untargeted} Loss \sbcode{with} Z $\mid$ \\
&  \sbcode{targeted} Loss,\;n\;-\;\sbcode{untargeted}\;Loss\;\sbcode{with}\;Z \\[0.3em]
Z ::= & logits $\mid$ probs \\[0.3em]
Loss ::= & CrossEntropy $\mid$ HingeLoss $\mid$ L1 $\mid$
  DLR $\mid$ LogitMatching \\
\end{tabular}
}



\emph{Targeted vs Untargeted}. The loss can be either untargeted, where the goal is to change the classification to any other label $f(x') \neq f(x)$, or targeted, where the goal is to predict a concrete label $f(x') = l$. Even though the untargeted loss is less restrictive, it is not always easier to optimize in practice, and replacing it with a targeted attack might perform better. 
When using \sbcode{targeted}\scode{ Loss, n}, the attack will consider the top \scode{n} classes with the highest probability as the targets.

\emph{Loss Formulation}. The concrete loss formulation includes loss functions used in existing adaptive attacks, as well as the recently proposed difference in logit ratio loss~\citep{croce2020reliable}. We provide a formal definition of the loss functions used in our work in Appendix \ref{sec_search_space}.

\emph{Logits vs. Probabilities}. In our search space, loss functions can be instantiated both with logits as well as with probabilities. Note that some loss functions are specifically designed for one of the two options, such as \scode{C\&W}~\citep{CandW} or \scode{DLR}~\citep{croce2020reliable}. 
While such knowledge can be used to reduce the search space, it is not necessary as long as the search algorithm is powerful enough to recognize that such a combination leads to poor results.

\emph{Loss Replacement}. Because the key idea behind many of the defenses is finding a property that helps differentiate between adversarial and natural images, one can also define the optimization objective in the same way. These feature-level attacks~\citep{sabour2015adversarial} avoid the need to directly optimize the complex objective defined by the adversarial defense and have been effective at circumventing them.
As an example, the logit matching loss minimizes the difference of logits between adversarial sample $x'$ and a natural sample of the target class~$x$ (selected at random from the dataset). Instead of logits, the same idea can also be applied to other statistics, such as internal representations computed by a pre-trained model or KL-divergence between label probabilities.

\section{Search Algorithm}\label{search}
We now describe our search algorithm that optimizes the problem statement from \Eqref{problem_statement}. Since we do not have access to the underlying distribution, we approximate \Eqref{problem_statement} using the dataset $\dataset$ as follows:
\begin{equation}\textstyle
score(f, a, \dataset) = \frac{1}{\abs{\dataset}} \sum_{i = 1}^{\abs{\dataset}} - \lambda l_a + \max_{d(x', x) \leq \epsilon} c(f, a(x, f), x)
\end{equation}
where $a \in \gA$ is an attack, $l_a \in \sR^+$ denotes untargeted cross-entropy loss of $a$ on the input, and $\lambda \in \sR$ is a hyperparameter.
The intuition behind $-\lambda \cdot l_a$ is that it acts as a tie-breaker in case the criterion $c$ alone is not enough to differentiate between multiple attacks.
While this is unlikely to happen when evaluating on large datasets, it is quite common when using only a small number of samples.
Obtaining good estimates in such cases is especially important for achieving scalability since performing the search directly on the full dataset would be prohibitively slow.

\setlength{\textfloatsep}{15pt}
\begin{algorithm*}[t]

\caption{A search algorithm that given a model $f$ with unknown defense, discovers an adaptive attack from the attack search space $\gA$ with the best $score$.}\label{SearchAlgo}

\SetKwFunction{Search}{AdaptiveAttackSearch}
\SetKwRepeat{Do}{do}{while}%
\SetKwProg{myproc}{def}{}{}
\nonl \myproc{\Search}{
\KwIn{dataset $\dataset$, model $f$, attack search space $\gA= \sS \times \sT$, number of trials $k$, initial dataset size~$n$,  attack sequence length $m$, criterion function $c$, initial parameter estimator model $M$, default attack~$\Delta \in \sS$}
\KwOut{adaptive attack from $a_{[s, t]}\in \gA = \sS \times \sT$ achieving the highest $score$ on $\dataset$ }
{$t \gets \argmax_{t \in \sT} score(f, a_{[\Delta, t]}, \dataset)$ \algorithmiccomment{Find surrogate model $t$ using default attack $\Delta$}\\}
{$\gS \gets \bot$ \algorithmiccomment{Initialize attack to be no attack, which returns the input image}\\}
\For(\algorithmiccomment{Run $m$ iterations to get sequence of $m$ attacks}){$j \gets 1\!:\!m$}{
{$\dataset \gets \dataset \setminus \{ x \mid x \in \dataset \wedge c(f,a_{[\gS,t]}(x,f), x)\} $ \algorithmiccomment{Remove non-robust samples}\\}
{$\gH \gets \emptyset;~\dataset_{\texttt{sample}} \gets sample(\dataset, n)$\algorithmiccomment{Initial dataset with $n$ samples}\\}
\For(\algorithmiccomment{Select candidate adaptive attacks}){$i \gets 1\!:\!k$}{
	$\theta' \gets \argmax_{\theta \in \sS} P(\theta \mid M)$ \algorithmiccomment{Best unseen parameters according to the model $M$}\\
	$q \gets score(f, a_{[\theta',t]}, \dataset_\texttt{sample})$\\
	$\gH \gets \gH \cup \{(\theta', q)\}$\\
	$M \gets$ update model $M$ with $(\theta', q)$\\
}
$\gH \gets $ keep $\abs{\gH} / 4$ attacks with the best score\\
\While(\algorithmiccomment{Successive halving (SHA)}){$ \abs{\gH} > 1 ~\textbf{and}~ \dataset_\texttt{sample} \neq \dataset$}{
$\dataset_\texttt{sample} \gets \dataset_\texttt{sample} \cup sample(\dataset \setminus \dataset_\texttt{sample}, \abs{\dataset_\texttt{sample}})$\algorithmiccomment{Double the dataset size}\\
$\gH \gets \{ (\theta, score(f, a_{[\theta, t]}, \dataset_\texttt{sample})) \mid (\theta, q) \in \gH \}$\algorithmiccomment{Re-evaluate on larger dataset}\\
$\gH \gets $ keep $\abs{\gH} / 4$ attacks with the best score\\
}
$\gS \gets \gS;~$best~attack~in~$\gH$
}
\Return{ $a_{[\gS, t]}$ }
}
\vspace{-0.2em}
\end{algorithm*} 
\paragraph{Search Algorithm}
We present our search algorithm in \Algref{SearchAlgo}.
We start by searching through the space of network transformations $t\!\in\!\sT$ to find a suitable surrogate model (line~1). This is achieved by taking the default attack~$\Delta$ (in our implementation, we set $\Delta$ to \apgdce{}), and then evaluating all locations where \scode{BPDA} can be used, and subsequently evaluating all layers that can be removed. Even though this step is exhaustive, it takes only a fraction of the runtime in our experiments, and no further optimization was necessary.

Next, we search through the space of attacks $\sS$. As this search space is enormous, we employ three techniques to improve scalability and attack quality. First, to generate a sequence of $m$ attacks, we perform a greedy search (lines 3-16). That is, in each step, we find an attack with the best score on the samples not circumvented by any of the previous attacks (line~4).
Second, we use a parameter estimator model $M$ to select the suitable parameters (line 7). In our work, we use Tree of Parzen Estimators~\citep{bergstra2011algorithms}, but the concrete implementation can vary. Once the parameters are selected, they are evaluated using the $score$ function (line 8), the result is stored in the trial history $\gH$ (line 9), and the estimator is updated (line 10). 
Third, because evaluating the adversarial attacks can be expensive, and the dataset $\dataset$ is typically large, we employ successive halving technique~\citep{karnin13, jamieson16}.
Concretely, instead of evaluating all the trials on the full dataset, we start by evaluating them only on a subset of samples $\dataset_{\texttt{sample}}$ (line 5). Then, we improve the score estimates by iteratively increasing the dataset size (line 13), re-evaluating the scores (line 14), and retaining a quarter of the trials with the best score (line 15). We repeat this process to find a~single best attack from $\gH$, which is then added to the sequence of attacks $\gS$ (line 16). 

\paragraph{Time Budget and Worst-case Search Time} 
We set a time budget on the attacks, measured in seconds per sample per attack, to restrict resource-expensive attacks and allow the tradeoff between computation time and attack strength. If an attack exceeds the time limit in line 8, the evaluation terminates, and the score is set to be $-\infty$. We analyzed the worst-case search time to be $4/3\times$ the allowed attack runtime in our experiments, which means the search overhead is both controllable and reasonable in practice. The derivation is shown in Appendix \ref{appendix_time_complexity}.




\section{Evaluation}\label{sec_eval}

We evaluate \tool{} on 24 models with diverse defenses and compare the results to \scode{AutoAttack}~\citep{croce2020reliable} and to several existing handcrafted attacks. \scode{AutoAttack} is a state-of-the-art tool designed for reliable evaluation of adversarial defenses that improved the originally reported results for many existing defenses by up to 10\%. Our key result is that \tool finds stronger or similar attacks than \scode{AutoAttack} for virtually all defenses:

\begin{itemize}[leftmargin=*]
\item In 10 cases, the attacks found by \tool are significantly stronger than \scode{AutoAttack}, resulting in 3.0\% to 50.8\% additional adversarial examples.

\item In the other 14 cases, \tool's attacks are typically 2x faster while enjoying similar attack quality.
\end{itemize}



\paragraph{Model Selection} We selected 24 models and divided them into three blocks \scode{A}, \scode{B}, \scode{C} as listed in \Tabref{eval_full}. Block \scode{A} contains diverse defenses with $\epsilon=4/255$. Block \scode{B} contains selected top models from \scode{RobustBench} ~\citep{croce2020robustbench}. Block \scode{C} contains diverse defenses with $\epsilon=8/255$. 

\paragraph{The \tool tool} The implementation of \tool is based on \scode{PyTorch}~\citep{PyTorch}, the implementations of \scode{FGSM}, \scode{PGD}, \scode{NES}, and \scode{DeepFool} are based on \scode{FoolBox}~\citep{rauber2017foolbox} version 3.0.0, \scode{C\&W} is based on \scode{ART}~\citep{DBLP:journals/corr/abs-1807-01069} version 1.3.0, and the attacks  \scode{APGD}, \scode{FAB}, and \scode{SQR} are from \citep{croce2020reliable}.
We use \scode{AutoAttack}'s \textit{rand} version if a defense has a randomization component, and otherwise we use its \textit{standard} version.
To allow for a fair comparison, we extended \scode{AutoAttack} with backward pass differential approximation (\scode{BPDA}), so we can run it on defenses with non-differentiable components; without this, all gradient-based attacks would fail.
We instantiate \Algref{SearchAlgo} by setting: the attack sequence length  $m=3$, the number of trials $k=64$, the initial dataset size $n=100$, and we use a time budget of $0.5$ to $3$ seconds per sample depending on the model size.
All of the experiments are performed using a single RTX 2080 Ti GPU.

\begin{table*}[t]
\tra{1.15}
\centering
\caption{Comparison of \scode{AutoAttack} (\scode{AA}) and our approach (\tool{}) on 24 defenses. Further details of each defense, discovered adaptive attacks and confidence intervals are included in Appendix \ref{appendix_discovered_att} and \ref{conf_interval}.}
\label{eval_full}
\vskip 0.05in
{
\footnotesize

\begin{tabular}{@{}p{0.5cm}p{3.0cm}ccc c ccc c c@{}}


&& \multicolumn{3}{c}{Robust Accuracy (1 - Rerr)} & &\multicolumn{3}{c}{Runtime (min)} &  & Search \\[-0.2em] \cmidrule{3-5} \cmidrule{7-9} \cmidrule{11-11}
\multicolumn{2}{l}{\fontfamily{lmtt}\selectfont \textbf{CIFAR-10}, $l_\infty$, $\epsilon=4/255$} & \scode{AA} & \tool{} & $\quad \Delta \quad$ & & \scode{AA} & \tool{} & Speed-up & & \tool{} \\[-0.2em]
\midrule

\scode{A1} & \citet{madry2017towards}  & 44.78 & \textbf{ 44.69 } & { \cellcolor{green!30} -0.09 } & & 25 & 20 & { \cellcolor{green!30} 1.25$\times$ } & & 88\\
\scode{A2$^{\dagger}$} & \citet{buckman2018thermometer} & \;\;2.29 & \textbf{ \;\;1.96 } & { \cellcolor{green!30} -0.33 } & & 9 & 7 & { \cellcolor{green!30} 1.29$\times$ } & & 116\\
\scode{A3$^{\dagger}$} & \citet{das2017keeping} + \citet{lee2018defending} & \;\;0.59 & \textbf{ \;\;0.11 } & { \cellcolor{green!30} -0.48 } & & 6 & 2 & { \cellcolor{green!30} 3.00$\times$ } & & 40\\
\scode{A4} & \citet{metzen2017detecting} & \;\;6.17 & \textbf{ \;\;3.04 } & { \cellcolor{green!30} -3.13 } & & 21 & 13 & { \cellcolor{green!30} 1.62$\times$ } & & 80\\
\scode{A5} & \citet{guo2018countering} & 22.30 & \textbf{ 12.14 } & { \cellcolor{green!30} -10.16 } & & 19 & 17 & { \cellcolor{green!30} 1.12$\times$ } & & 99\\
\scode{A6$^{\dagger}$} & \citet{pang2019improving} & \;\;4.14 & \textbf{ \;\;3.94 } & { \cellcolor{green!30} -0.20 } & & 28 & 24 & { \cellcolor{green!30} 1.17$\times$ } & & 237\\
\scode{A7} & \citet{DBLP:journals/corr/PapernotMWJS15} & \;\;2.85 & \textbf{ \;\;2.71 } & { \cellcolor{green!30} -0.14 } & & 4 & 4 & {  1.00$\times$ } & & 84\\
\scode{A8} & \citet{Xiao2020Enhancing} & 19.82 & \textbf{ 11.11 } & { \cellcolor{green!30} -8.71 } & & 49 & 22 & { \cellcolor{green!30} 2.23$\times$ } & & 189\\
\scode{A9} & \citet{Xiao2020Enhancing}$_\texttt{ADV}$ & 64.91 & \textbf{ 63.56 } & { \cellcolor{green!30} -1.35 } & & 157 & 100 & { \cellcolor{green!30} 1.57$\times$ } & & 179\\
\scode{A9'} & \citet{Xiao2020Enhancing}$_\texttt{ADV}$ & 64.91 & \textbf{ 17.70 } & { \cellcolor{green!30} -47.21 } & & 157 & 2,280 & {  0.07$\times$ } & & 1,548\\

\\[-0.6em]
\multicolumn{2}{l}{\fontfamily{lmtt}\selectfont \textbf{CIFAR-10}, $l_\infty$, $\epsilon=8/255$} &\\
\midrule
\scode{B10$^*$} & \citet{gowal2021uncovering} & 62.80 & \textbf{ 62.79 } & { \cellcolor{green!30} -0.01 } & & 818 & 226 & { \cellcolor{green!30} 3.62$\times$ } & & 761\\
\scode{B11$^*$} & \citet{wu2020adversarial}$_\texttt{RTS}$ & 60.04 & \textbf{ 60.01 } & { \cellcolor{green!30} -0.03 } & & 706 & 255 & { \cellcolor{green!30} 2.77$\times$ } & & 690\\
\scode{B12$^*$} & \citet{zhang2021geometryaware} & 59.64 & \textbf{ 59.56 } & { \cellcolor{green!30} -0.08 } & & 604 & 261 & { \cellcolor{green!30} 2.31$\times$ } & & 565\\
\scode{B13$^*$} & \citet{NEURIPS2019_32e0bd14} & 59.53 & \textbf{ 59.51 } & { \cellcolor{green!30} -0.02 } & & 638 & 282 & { \cellcolor{green!30} 2.26$\times$ } & & 575\\
\scode{B14$^*$} & \citet{sehwag2020hydra} & \textbf{ 57.14 } & 57.16 & { \cellcolor{red!30} 0.02 } & & 671 & 429 & { \cellcolor{green!30} 1.56$\times$ } & & 691\\

\\[-0.6em]
\multicolumn{2}{l}{\fontfamily{lmtt}\selectfont \textbf{CIFAR-10}, $l_\infty$, $\epsilon=8/255$} &\\
\midrule
\scode{C15$^*$} & \citet{stutz2020confidencecalibrated} & 77.64 & \textbf{ 39.54 } & { \cellcolor{green!30} -38.10 } & & 101 & 108 & {  0.94$\times$ } & & 296\\
\scode{C15'} & \citet{stutz2020confidencecalibrated} & 77.64 & \textbf{ 26.87 } & { \cellcolor{green!30} -50.77 } & & 101 & 205 & {  0.49$\times$ } & & 659\\
\scode{C16$^*$} & \citet{feature_scatter} & \textbf{ 36.74 } & 37.11 & { \cellcolor{red!30} 0.37 } & & 381 & 302 & { \cellcolor{green!30} 1.26$\times$ } & & 726\\
\scode{C17} & \citet{grathwohl2020classifier} & \textbf{ \;\;5.15 } & \;\;5.16 & { \cellcolor{red!30} 0.01 } & & 107 & 114 & { \cellcolor{red!30}   0.94$\times$ } & & 749\\
\scode{C18} & \citet{Xiao2020Enhancing}$_\texttt{ADV}$ & \;\;5.40 & \textbf{ \;\;2.31 } & { \cellcolor{green!30} -3.09 } & & 95 & 146 & {  0.65$\times$ } & & 828\\
\scode{C19} & \citet{wang2019resnets} & 50.84 & \textbf{ 50.81 } & { \cellcolor{green!30} -0.03 } & & 734 & 372 & { \cellcolor{green!30} 1.97$\times$ } & & 755\\
\scode{C20$^{\dagger}$} & \scode{B11} + Defense in \scode{A3} & 60.72 & \textbf{ 60.04 } & { \cellcolor{green!30} -0.68 } & & 621 & 210 & { \cellcolor{green!30} 2.96$\times$ } & & 585\\
\scode{C21$^{\dagger}$} & \scode{C17} + Defense in \scode{A3} & 15.27 & \textbf{ \;\;5.24 } & { \cellcolor{green!30} -10.03 } & & 261 & 79 & { \cellcolor{green!30} 3.30$\times$ } & & 746\\
\scode{C22} & \scode{B11} + Random Rotation & 49.53 & \textbf{ 41.99 } & { \cellcolor{green!30} -7.54 } & & 255 & 462 & {  0.55$\times$ } & & 900\\
\scode{C23} & \scode{C17} + Random Rotation & 22.29 & \textbf{ 13.45 } & { \cellcolor{green!30} -8.84 } & & 114 & 374 & {  0.30$\times$ } & & 1,023\\
\scode{C24} & \citet{Yu_2019_NeurIPSweakness} & \;\;6.25 & \textbf{ \;\;3.07 } & { \cellcolor{green!30} -3.18 } & & 110 & 56 & { \cellcolor{green!30} 1.96$\times$ } & & 502\\

\bottomrule
\\[-1.1em]
\multicolumn{11}{l}{$^*$model available from the authors, $^{\dagger}$model with non-differentiable components. } \\
\multicolumn{11}{l}{\scode{B12} uses $\epsilon=0.031$. \scode{C15} uses $\epsilon=0.03$. \scode{A9'} uses time budget $T_{c}=30$. \scode{C15'} uses $m=8$.}\\

\end{tabular}
}
\vskip -0.15in
\end{table*}

\paragraph{Evaluation Metric}
Following \citet{stutz2020confidencecalibrated}, we use the {\em robust test error (Rerr)} metric to combine the evaluation of defenses with and without detectors. We include details in Appendix \ref{appendix_eval_details}. In our evaluation, \tool{} produces consistent results on the same model across independent runs with the standard deviation $\sigma < 0.2$ (computed across 3 runs). The details are included in Appendix \ref{conf_interval}.


\paragraph{Comparison to \scode{AutoAttack}}
Our main results, summarized in \Tabref{eval_full}, show the robust accuracy (lower is better) and runtime of both \scode{AutoAttack} (\autoattack{}) and \tool{} over the 24 defenses.
For example, for \scode{A8} our tool finds an attack that leads to lower robust accuracy (11.1\% for \tool vs. 19.8\% for \autoattack{}) and is more than twice as fast (22 min for \tool vs. 49 min for \autoattack).
Overall, \tool significantly improves upon \autoattack{} or provides similar but faster attacks.

We note that the attacks from \autoattack{} are included in our search space (although without the knowledge of their best parameters and sequence), and so it is expected that \tool{} performs at least as well as \autoattack{}, provided sufficient exploration time.
Importantly, \tool{}~often finds better attacks: for 10 defenses, \tool{} reduces the~robust accuracy by 3\% to 50\% compared to \autoattack{}.
Next, we discuss the results in more~detail.

\textit{Defenses based on Adversarial Training.} Models in block \scode{B} are selected from \scode{RobustBench} \citep{croce2020robustbench}, and they are based on various extensions of adversarial training, such as using additional unlabelled data for training, extensive hyperparameter tuning, instance weighting or loss regularization. 
The results show that the robustness reported by \scode{AA} is already very high and using \tool{} leads to only marginal improvement. However, because our tool also optimizes for the runtime, \tool{} does achieve significant speed-ups, ranging from 1.5$\times$  to 3.6$\times$.
The reasons behind the marginal robustness improvement of \tool{} are two-fold. First, it shows that \tool{} is limited by the attack techniques search space, as the attack found are all variations of \scode{APGD}. 
Second, the models \scode{B10} - \scode{B14} aim to improve the adversarial training procedure rather than developing a new defence. This is in contrast to models that do design various types of new defences (included in blocks \scode{A} and \scode{C}), evaluating which typically requires discovering a new adaptive attack. For these new defences, evaluation is much more difficult and this is where our approach also improves the most.

\textit{Obfuscation Defenses.} Defenses \scode{A3}, \scode{A8}, \scode{A9}, \scode{C18}, \scode{C20}, and \scode{C21} are based on gradient obfuscation. \tool{} discovers stronger attacks that reduce the robust accuracy for all defenses by up to 47.21\%.
Here, removing the obfuscated defenses in \scode{A3}, \scode{C20}, and \scode{C21} provides better gradient estimation for the attacks. Further, the use of more suitable loss functions strengthens the discovered attacks and improves the evaluation results for \scode{A8} and \scode{C18}.

\textit{Randomized Defenses.} For the randomized input defenses \scode{A8}, \scode{C22}, and \scode{C23}, \tool{} discovers attacks that, compared to \autoattack{}'s \textit{rand} version, further reduce robustness by 8.71\%, 7.54\%, and 8.84\%, respectively. This is achieved by using stronger yet more costly parameter settings, attacks with different backbones (\scode{APGD}, \scode{PGD}) and 7 different loss functions (as listed in Appendix \ref{appendix_discovered_att}).

\textit{Detector based Defenses.} For \scode{C15}, \scode{A4}, and \scode{C24} defended with detectors, \tool{} improves over \autoattack{} by reducing the robustness by 50.77\%, 3.13\%, and 3.18\%, respectively.
This is because none of the attacks discovered by \tool{} are included in \autoattack{}.
Namely, \tool{} found \scode{SQR}$_{\texttt{DLR}}$ and \scode{APGD}$_{\texttt{Hinge}}$ for \scode{C15}, untargeted \scode{FAB} for \scode{A4} (\scode{FAB} in \autoattack{} is targeted), and \scode{PGD}$_{\texttt{L1}}$ for~\scode{C24}.

\paragraph{Generalization of \tool{}}
Given a new defense, the main strength of our approach is that it directly benefits from all existing techniques included in the search space. 
Here, we compare our approach to three handcrafted adaptive attacks not included in the search space.

As a first example, \scode{C15}~\citep{stutz2020confidencecalibrated} proposes an adaptive attack \scode{PGD-Conf} with backtracking that leads to robust accuracy of 36.9\%, which can be improved to~31.6\% by combining \scode{PGD-Conf} with blackbox attacks. \tool{} finds \scode{APGD}$_{\texttt{Hinge}}$ and \scode{Z}\;=\;\scode{probs}. This combination is interesting since~the hinge loss maximizing the difference between the top two predictions, in fact, reflects the \scode{PGD-Conf} objective function. Further, similarly to the manually crafted attack by \scode{C15}, a different blackbox attack included in our search space, \scode{SQR}$_{\texttt{DLR}}$, is found to complement the strength of \scode{APGD}. When using a sequence of three attacks, we achieve 39.54\% robust accuracy. We can decrease the robust accuracy even further by increasing the number of attacks to eight -- the robust accuracy drops to 26.87\%, which is a stronger result than the one reported in the original paper. In this case, our search space and the search algorithm are powerful enough to not only replicate the main ideas of \citet{stutz2020confidencecalibrated} but also to improve its evaluation when allowing for a larger attack budget. Note that this improvement is possible even without including the backtracking used by \scode{PGD-Conf} as a building block in our search space. In comparison, the robust accuracy reported by \autoattack{} is only 77.64\%.

As a second example, \scode{C18} is known to be susceptible to \scode{NES} which achieves 0.16\% robust accuracy \citep{tramer2020adaptive}.
To assess the quality of our approach, we remove \scode{NES} from our search space and instead try to discover an adaptive attack using the remaining building blocks. 
In this case, our search space was expressive enough to find an alternative attack that achieves 2.31\% robust accuracy. 

As a third example, to break \scode{C24}, \citet{tramer2020adaptive} designed an adaptive attack that linearly interpolates between the original and the adversarial samples using \scode{PGD}. This technique breaks the defense and achieves 0\% robust accuracy. In comparison, we find \scode{PGD}$_{\texttt{L1}}$, which achieves 3.07\% robust accuracy.
In this case, the fact that \scode{PGD}$_{\texttt{L1}}$ is a relatively weak attack is an advantage -- it successfully bypasses the detector by not generating overconfident predictions.

\paragraph{\tool{} Interpretability} 
As illustrated above, it is possible to manually analyze the discovered attacks in order to
understand how they break the defense mechanism. Further, we can also gain insights from the patterns of attacks searched across all the models (shown in Appendix \ref{appendix_discovered_att},  \Tabref{result_complete}). For example, it turns out that $\ell_{\texttt{CE}}$ is not as frequent as $\ell_{\texttt{DLR}}$ or $\ell_{\texttt{hinge}}$. This fact challenges the common practice of using $\ell_{\texttt{CE}}$ as the default loss when evaluating robustness. In addition, using $\ell_{\texttt{CE}}$ during adversarial training can make models resilient to $\ell_{\texttt{CE}}$, loss, but not necessarily to other losses. 
 

%
%

\paragraph{\tool{} Scalability}
To assess \tool{}'s scalability, we perform two ablation studies: \captionn{i} increase the search space by $4\times$ (by adding $8$ random attacks, their corresponding parameters, and $4$ dummy losses), and \captionn{ii} keep the search space size unchanged but reduce the search runtime by half.
In \captionn{i}, we observed a marginal performance decrease when using the same runtime, and we can reach the same attack strength when the runtime budget is increased by $1.5\times$.
In \captionn{ii}, even when we reduce the runtime by half, we can still find attacks that are only slightly worse ($\leq 0.4$). This shows that a budget version of the search can provide a strong robustness evaluation. We include detailed results in Appendix~\ref{appendix_scalability}.



\paragraph{Ablation Studies}
Similar to existing handcrafted adaptive attacks, all three components included in the search space were important for generating strong adaptive attacks for a variety of defenses.
Here we briefly discuss their importance while including the full experiment results in Appendix \ref{appendix_ablation}.

\textit{Attack \& Parameters.} We demonstrate the importance of parameters by comparing \scode{PGD}, \scode{C\&W}, \scode{DF}, and \scode{FGSM} with default library parameters to the best configuration found when available parameters are included in the search space.
The attacks found by \tool{} are on average 5.5\% stronger than the best attack among the four attacks on \scode{A} models.

\textit{Loss Formulation.} To evaluate the effect of modeling different loss functions, we remove them from the search space and keep only the original loss function defined for each attack. The search score drops by 3\% on average for \scode{A} models without the loss formulation.

\begin{wraptable}{r}{6.6cm}
\caption{The robust accuracy (1 - Rerr) of networks with different \scode{BPDA} policies evaluated by \apgdce{} with 50 iterations.}
    \label{BPDA_comparison}
    \begin{center}
    {
\footnotesize
\tra{1.15}
    \begin{tabular}{@{}rccccccc@{}}
    \toprule
\scode{BPDA Type} & \scode{A2} & \scode{A3} & \scode{C20} & \scode{C21}\\
\midrule
identity 			& 18.5 & \textbf{9.6} & \textbf{70.5} & \textbf{84.0}\\
1x1 convolution 	& 8.9 & 10.3 & 70.8 & 84.9\\
2 layer conv+ReLU & \textbf{3.7} & 14.9 & 74.1 & 86.2\\
\bottomrule
\end{tabular}
}
\end{center}
\end{wraptable} 
\textit{Network Processing.} 
In \scode{C21}, the main reason for achieving 10\% decrease in robust accuracy is the removal of the gradient obfuscated defense Reverse Sigmoid. We provide a more detailed ablation in \Tabref{BPDA_comparison}, which shows the effect of different \scode{BPDA} instantiations included in our search space. For \scode{A2}, since the non-differentiable layer is non-linear thermometer encoding, it is better to use a function with non-linear activation to approximate it. For \scode{A3}, \scode{C20}, \scode{C21}, the defense is image \scode{JPEG} compression and identity network is the best algorithm since the networks can overfit when training on limited data.




\section{Related Work}

The most closely related work to ours is \scode{AutoAttack} \citep{croce2020reliable}, which improves the evaluation of adversarial defenses by proposing an ensemble of four fixed attacks. Further, the key to stronger attacks was a new algorithm \scode{APGD}, which improves upon \scode{PGD} by halving the step size dynamically based on the loss at each step.
In our work, we improve over \scode{AutoAttack} in three keys aspects: \captionn{i} we formalize a search space of adaptive attacks, rather than using a fixed ensemble, \captionn{ii} we design a search algorithm that discovers the best adaptive attacks automatically, significantly improving over the results of \scode{AutoAttack}, and \captionn{iii} our search space is extensible and allows reusing building blocks from one attack by other attacks, effectively expressing new attack instantiations. For example, the idea of dynamically adapting the step size is not tied to \scode{APGD}, but it is a general concept applicable to any step-based algorithm.

Another related work is Composite Adversarial Attacks (\scode{CAA}) \citep{mao2020composite}. The main idea of \scode{CAA} is that instead of selecting an ensemble of four attacks that complement each other as done by \scode{AutoAttack}, the authors proposed to search for a sequence of attacks that achieve the best performance.
Here, the authors focus on evaluating defences based on adversarial training and show improvements of up to 1\% over \scode{AutoAttack}.
In comparison, our main idea is that the way adaptive attacks are designed today can be formalized as a search space that includes not only sequence of attacks but also loss functions, network processing and rich space of hyperparameters. This is critical as it defines a much larger search space to cover a wide range of defenses, beyond the reach of both \scode{CA} and \scode{AutoAttack}. This can be also seen in our evaluation -- we achieve significant improvement by finding 3\% to 50\% more adversarial examples for 10 models.


Our work is also closely related to the recent advances in AutoML, such as in the domain of neural architecture search (NAS)~\citep{zoph2017neural, elsken2019neural}.
Similar to our work, the core challenge in NAS is an efficient search over a large space of parameters and configurations, and therefore many techniques can also be applied to our setting. This includes BOHB \citep{falkner2018bohb}, ASHA \citep{li2018massively}, using gradient information coupled with reinforcement learning~\citep{zoph2017neural} or continuous search space formulation~\citep{liu2019darts}.
Even though finding completely novel neural architectures is often beyond the reach, NAS is still very useful and finds many state-of-the-art models.
This is also true in our setting -- while human experts will continue to play a key role in defining new types of adaptive attacks, as we show in our work, it is already possible to automate many of the intermediate steps. 
\section{Conclusion}

We presented the first tool that aims to automate the process of finding strong adaptive attacks specifically tailored to a given adversarial defense. Our key insight is that we can identify reusable techniques used in existing attacks and formalize them into a search space. Then, we can phrase the challenge of finding new attacks as an optimization problem of finding the strongest attack over this search space.

Our approach automates the tedious and time-consuming trial-and-error steps that domain experts perform manually today, allowing them to focus on the creative task of designing new attacks.
By doing so, we also immediately provide a more reliable evaluation of new and existing defenses, many of which have been broken only after their proposal because the authors struggled to find an effective attack by manually exploring the vast space of techniques.
Importantly, even though our current search space contains only a subset of existing techniques, our evaluation shows that \tool{} can partially re-discover or even improve upon some handcrafted adaptive attacks not included in our search~space. 

However, there are also limitations to overcome in future work. First, while the search space can be easily extended, it is also inherently incomplete, and domain experts will still play an important role in designing novel types of attacks. Second, the search algorithm does not model the attack runtime and as a result, incorporating expensive attacks can be computational unaffordable. This is problematic as it can incur huge overhead even if a fast attack does exist. Finally, an interesting future work is to use meta-learning to improve the search even further, allowing \tool{} to learn across multiple models, rather than starting each time from scratch.


\section{Societal Impacts}
In this paper, an approach to improve the evaluation of adversarial defenses by automatically finding adaptive adversarial attacks is proposed and evaluated.
As such, this work builds on a large body of existing research on developing adversarial attacks and defences and thus shares the same societal impacts.
Concretely, the presented approach can be used both in a beneficial way by the researchers developing adversarial defenses, as well as, in a malicious way by an attacker trying to break existing models. In both cases, the approach is designed to improve empirical model evaluation, rather than providing verified model robustness, and thus is not intended to provide formal robustness guarantees for safety-critical applications. 
For applications where formal robustness guarantess are required, instead of using empirical techniques as in this work, one should instead adapt the concurrent line of work on certified robustness.

\bibliography{bib}
\bibliographystyle{icml2021}
\cleardoublepage

\appendix

\section{\tool{} Time Complexity}\label{appendix_time_complexity}

This section gives the worst-case time analysis for \Algref{SearchAlgo}. We denote $T_{a}$ to be the attack time and $T_{r}$ to be the search time. We will show that with the per sample per attack time constraint of $T_{c}$: 
\begin{equation}
\label{attack_time_bound}
T_{a} \le m \times N \times T_{c} 
\end{equation}
\begin{equation}
\label{search_time_bound}
T_{r} \le 2 \times m \times n \times k \times T_{c} 
\end{equation} 
Where $m$, $N$, $n$, $k$ are the number of attacks, the size of the dataset $D$, the size of initial dataset size, the number of attacks to sample respectively.

In \Algref{SearchAlgo}, only steps on lines 1,4,8,14 are timing critical as they apply the expensive attack algorithms. Other steps like sampling datasets and applying parameter estimator $M$ are considered as constant overhead. $T_{a}$ is the total runtime of line 4, because line 4 is the step to apply the attack on all the samples. $T_{r}$ includes the runtime of lines 1,8,14. 

$T_{a}$ has the worst-case runtime when each of the $m$ attacks uses the full time budget $T_{c}$ on all the samples (denoted as $N$). This gives the bound shown in \Eqref{attack_time_bound}. 


For $T_{r}$, we first analyze the time in lines 8 and 14 for a single attack. In line~8, the maximum time to perform $k$ attacks on $n$ samples is: $n \times k \times T_{c}$. In line 14, the cost of the first iteration is: $\frac{1}{2} n \times k \times T_{c}$ as there are $k/4$ attacks and $2n$ samples. The cost of SHA iteration is halved for every subsequent iteration by such design, so the total time for line 14 is $n \times k \times T_{c}$. As there are $m$ attacks, the total time bound for lines 8 and 14 is: $2 \times m \times n \times k \times T_{c}$. 

The runtime for line 1 is bounded by $N \times T_{fast}$ as we run single attack on all the samples. Here, we use $T_{fast}$ to denote the maximum runtime of a fast attack that we run at this stage. This step is typically negligible compared to the subsequent search, i.e., $N \times T_{fast} \ll 2 \times m \times n \times k \times T_{c}$. Overall, we can therefore bound the search runtime by considering the lines and 8 and 14, which leads to the bound from \Eqref{search_time_bound}. 


In our evaluation, we use $m=3, k=64, n=100, N=10000$. Substituting into \Eqref{search_time_bound} leads to $T_{r} \le 2 \times 3 \times 100 \times 64 \times T_{c} \le 4 \times N \times T_{c}$. This means the total search time is bounded by the time bound of executing a sequence of $4$ attacks on the full dataset. Further, $T_{r} \le \frac{4}{3} \times m \times N \times T_{c}$, which means the search time of an attack is bounded by $\frac{4}{3}$ of the allowed runtime to execute the attack. 




\section{Search Space of $\sS \times \sL$} \label{sec_search_space}

\subsection{Loss function space $\sL$}


Recall that the loss function search space is defined as:

{
\fontfamily{lmtt}\selectfont

\begin{tabular}{@{}rll@{}}
\multicolumn{3}{l}{(Loss Function Search Space)} \\[0.3em]
$\sL$ ::= & 
\sbcode{targeted} Loss, n \sbcode{with} Z $\mid$ \\
& \sbcode{untargeted} Loss \sbcode{with} Z $\mid$ \\
&  \sbcode{targeted} Loss,\;n\;-\;\sbcode{untargeted}\;Loss\;\sbcode{with}\;Z \\[0.3em]
Z ::= & logits $\mid$ probs \\[0.3em]
\end{tabular}
}

To refer to different settings, we use the following notation:
\begin{itemize}
\item \scode{U}: for the \sbcode{untargeted} loss,
\item \scode{T}: for the \sbcode{targeted} loss,
\item \scode{D}: for the \sbcode{targeted} $-$ \sbcode{untargeted} loss
\item \scode{L}: for using \scode{logits}, and
\item \scode{P}: for using \scode{probs}
\end{itemize}

For example, we use \scode{DLR-U-L} to denote \sbcode{untargeted} \scode{DLR} loss with \scode{logits}.
The loss space used in our evaluation is shown in \Tabref{loss_modifier_table}. For hinge loss, we set $\kappa=-\infty$ in implementation to encourage stronger adversarial samples. Effectively, the search space includes all the possible combinations expect that the cross-entropy loss supports only probability. Note that although $\ell_{\texttt{DLR}}$ is designed for logits, and $\ell_{\texttt{LogitMatching}}$ is designed for targeted attacks, the search space still makes other possibilities an option (i.e., it is up to the search algorithm to learn which combinations are useful and which are not).

\begin{table}[t]
\caption{Loss functions and their modifiers. \ding{51} means the loss supports the modifier. \scode{P}~means the loss always uses Probability.}
\vspace{-1.3em}
\label{loss_modifier_table}
\begin{center}
  \begin{tabular}{ @{}lccccc@{} } 
  \toprule
	Name & Targeted & Logit/Prob & Loss \\
	\midrule
	$\ell_{\texttt{CE}}$ & \ding{51} & \scode{P} & \( \textstyle
\ell_{\texttt{CrossEntropy}} = - \sum_{i = 1}^{K} y_i \log(Z(x)_i) \)  \\[0.8em]
	$\ell_{\texttt{Hinge}}$ & \ding{51} & \ding{51} & \( \displaystyle \ell_{\texttt{HingeLoss}} = \max(- Z(x)_y + \max_{i \neq y} Z(x)_i, -\kappa)\) \\[0.8em]
	$\ell_{\texttt{L1}}$ & \ding{51} & \ding{51} & \( \displaystyle \ell_{\texttt{L1}} = - Z(x)_y \) \\[0.8em]
	$\ell_{\texttt{DLR}}$ & \ding{51} & \ding{51} & \(\displaystyle \ell_{\texttt{DLR}} = - \frac{Z(x)_y - \max_{i \neq y} Z(x)_i}{Z(x)_{\pi_1} - Z(x)_{\pi_3}}\) \\[0.8em]
	$\ell_{\texttt{LogitMatching}}$ & \ding{51} & \ding{51} & \( \displaystyle
\ell_{\texttt{LogitMatching}} = \norm{Z(x') - Z(x)}_2^2 \) \\
	
  \bottomrule
  \end{tabular}
\end{center}
\end{table}


\begin{table*}
\caption{Generic parameters and loss support for each attack in the search space. For the \scode{loss} column, "-" means the loss is from the library implementation, and \ding{51} means the attack supports all the loss functions defined in \Tabref{loss_modifier_table}. In other columns \ding{51} means the attack supports all the values, and the attack supports only the indicated set of values otherwise.}
\label{attack_loss_modifiers}
\begin{center}
{
\small
  \begin{sc}
  \begin{tabular}{ lcllcccr } 
  \toprule
    Attack & Randomize & EOT & Repeat & Loss & Targeted & logit/prob \\
  \midrule
	\scode{FGSM} &True& $\sZ[1,200]$& $^{*}\sZ[1, 10000]$ & \ding{51} & \ding{51} & \ding{51} \\
	\scode{PGD}  &True& $\sZ[1,40]$ & $\sZ[1,10]$ & \ding{51} & \ding{51} & \ding{51} \\
	\scode{DeepFool}   &False& $1$     & $1$ & \ding{51} & D & \ding{51} \\
	\scode{APGD} &True& $\sZ[1,40]$ & $\sZ[1,10]$ & \ding{51} & \ding{51} & \ding{51}  \\
	\scode{C\&W} &False& $1$      & $1$  & - & \{U, T\} & L \\
	\scode{FAB}  &True& $1$   & $\sZ[1,10]$ & - & \{U, T\} & L \\
	\scode{SQR}  &True& $1$    & $\sZ[1,3]$  & \ding{51} & \ding{51} & \ding{51} \\
	\scode{NES}  &True& $1$    & $1$   & \ding{51} & \ding{51} & \ding{51} \\
  \bottomrule
  \end{tabular}
  \end{sc}
  }
\end{center}

\end{table*}

\subsection{Attack Algorithm \& Parameters Space $\sS$}

Recall the attack space defined as:

{
\begin{tabular}{@{}llll@{}}
 $\qquad \sS$ ::= & $\sS$; $\sS$ $\mid$
\sbcode{randomize} $\sS$ $\mid$ \sbcode{EOT} $\sS$, n $\mid$ \sbcode{repeat} $\sS$, n $\mid$ \\ & \sbcode{try} $\sS$ \sbcode{for} n $\mid$ Attack \sbcode{with} params \sbcode{with} loss\;$\in \sL$ \\
\end{tabular}
}

\sbcode{randomize}, ~\sbcode{EOT}, ~\sbcode{repeat} are the generic parameters, and \scode{params} refers to attack specific parameters. The type of every parameter is either integer or float. An integer ranges from $p$ to $q$ inclusive is denoted as $\sZ[p,q]$. A float range from $p$ to $q$ inclusive is denoted as $\R[p,q]$. Besides value range, prior is needed for parameter estimator model (\scode{TPE} in our case), which is either uniform (default) or log uniform (denoted with $^{*}$). 
For example, $^{*}\sZ[1,100]$ means an integer value ranges from $1$ to $100$ with log uniform prior; $\R[0.1, 1]$ means a float value ranges from $0.1$ to $1$ with uniform prior. 

Generic parameters and the supported loss for each attack algorithm are defined in \Tabref{attack_loss_modifiers}. The algorithm returns a deterministic result if \sbcode{randomize} is False, and otherwise the results might differ due to randomization. 
Randomness can come from either perturbing the initial input or randomness in the attack algorithm.
Input perturbation is deterministic if the starting input is the original input or an input with fixed disturbance, and it is randomized if the starting input is chosen uniformly at random within the adversarial capability. For example, the first iteration of \scode{FAB} uses the original input but the subsequent inputs are randomized (if the randomization is enabled). Attack algorithms like \scode{SQR}, which is based on random search, has randomness in the algorithm itself. The deterministic version of such randomized algorithms is obtained by fixing the initial random seed.  

The definition of \sbcode{randomize} for \scode{FGSM}, \scode{PGD}, \scode{NES}, \scode{APGD}, \scode{FAB}, \scode{DeepFool}, \scode{C\&W} is whether to start from the original input or uniformly at random select a point within the adversarial capability. For \scode{SQR}, random means whether to fix the seed. We generally set \sbcode{randomize} to be True to allow repeating the attacks for stronger attack strength, yet we set \scode{DeepFool} and \scode{C\&W} to False as they are minimization attacks designed with the original inputs as the starting inputs. 

The attack specific parameters are listed in \Tabref{attack_specific_param}, and the ranges are chosen to be representative by setting reasonable upper and lower bounds to include the default values of parameters. Note that \scode{DeepFool} algorithm uses the loss \scode{D} to take difference between the predictions of two classes by design (i.e., \sbcode{targeted} $-$ \sbcode{untargeted} loss). \scode{FAB} uses loss similar to \scode{DeepFool}, and \scode{C\&W} uses the hinge loss. For \scode{C\&W} and \scode{FAB}, we just take the library implementation of the loss (i.e. without our loss function space formulation).  

\begin{table}
\caption{List of attack specific parameters. The parameter names correspond to the names in the library implementation}
\label{attack_specific_param}
\vspace{-1em}
\begin{center}

{
\small

\begin{minipage}{.45\linewidth}
  \begin{tabular}{ llr } 
  \toprule
    Attack & Parameter & Range and prior \\
  \midrule
    \multirow{3}{*}{\scode{NES}}& step & $\sZ[20, 80]$ \\
  & rel\_stepsize & $^{*}\R[0.01, 0.1]$ \\
  & n\_samples & $\sZ[400, 4000]$ \\
  \midrule
  \multirow{6}{*}{\scode{C\&W}} 
  & confidence & $\R[0,0.1]$ \\
  & max\_iter & $\sZ[20, 200]$ \\
  & binary\_search\_steps & $\sZ[5,25]$ \\
  & learning\_rate & $^{*}\R[0.0001,0.01]$ \\
  & max\_halving & $\sZ[5,15]$ \\
  & max\_doubling & $\sZ[5,15]$ \\
  \bottomrule
  \\
  \end{tabular}
  \end{minipage}
  \qquad \qquad 
\begin{minipage}{.45\linewidth}
 \begin{tabular}{ llr } 
  \toprule
 Attack & Parameter & Range and prior \\
  \midrule
 \multirow{2}{*}{\scode{PGD}} & step & $\sZ[20,200]$ \\ 
  & rel\_stepsize & $^{*}\R[1/1000, 1]$\\
  \midrule
  \multirow{2}{*}{\scode{APGD}} & rho & $\R[0.5, 0.9]$ \\
  & n\_iter & $\sZ[20, 500]$ \\
  \midrule
  \multirow{3}{*}{\scode{FAB}} 
  & n\_iter & $\sZ[10, 200]$ \\
  & eta & $\R[1, 1.2]$ \\
  & beta & $\R[0.7, 1]$ \\
  \midrule
  \multirow{2}{*}{\scode{SQR}} 
  & n\_queries & $\sZ[1000, 8000]$ \\
  & p\_init & $\R[0.5, 0.9]$ \\
  \bottomrule
  \end{tabular}
  \end{minipage}
}
\end{center}
\end{table}


\subsection{Search space conditioned on network property}
Properties of network defenses (e.g. randomized, detector, obfuscation) can be used as a prior to reduce the search space. In our work, \sbcode{EOT} is set to be $1$ for deterministic networks. Using meta-learning techniques to reduce the search space is left for future work.

\section{Evaluation Metrics Details} \label{appendix_eval_details}
We use the following $L_{\infty}$ criteria in the formulation:

\begin{figure}[h]
\begin{tikzpicture}

\node at (-1,0) {
\begin{minipage}{0.4\textwidth}
\begin{equation*}
\displaystyle
\norm{x' - x}_{\infty} \leq \epsilon \;\;\; \text{s.t.} \;\;\; \hat{f}(x') \neq \hat{f}(x)
\end{equation*}
\end{minipage}
};

\node[text width=2.8cm, align=center] at (-5.2,0) {\footnotesize
\textsc{Misclassification\\$L_{\infty}$ Attack}
};
\end{tikzpicture}
\vspace{-1.0em}
\end{figure}

We remove the misclassified clean input as a pre-processing step, such that the evaluation is performed only on the subset of correctly classified samples (i.e. $\hat{f}(x)=y$). 


\paragraph{Sequence of Attacks} Sequence of attacks defined in \Secref{search_space} is a way to calculate the per-example worst-case evaluation, and the four attack ensemble in AutoAttack is equivalent to sequence of four attacks [\scode{APGD}$_{\texttt{CE}}$, \scode{APGD}$_{\texttt{DLR}}$, \scode{FAB}, \scode{SQR}]. 
\Algref{seqattack} elaborates how the sequence of attacks is evaluated. 
That is, the attacks are performed in the order they were defined and the first sample $x'$ that satisfies the criterion $c$ is returned.

\begin{algorithm}[h!]
\label{seqattack}
\SetKwFunction{Search}{SeqAttack}
\SetKwRepeat{Do}{do}{while}%
\SetKwProg{myproc}{def}{}{}
\nonl \myproc{\Search}{
\KwIn{ model $f$, data $\vx$, sequence attacks $\gS \subseteq \sS$, network transformation $t \in \sT$, criterion function $c$ } 
\KwOut{ $\vx'$}
 \For{$\theta \in S$}{
  $x'$=$a_{[\theta, t]}(x, f)$\;
  \If{$c(f,x',x)$}{
  \Return{$x'$}
  }
 }
 \Return{$x'$}
}
 \caption{Sequence of attacks}
\end{algorithm}

\paragraph{Robust Test Error (Rerr)} 

Following \citet{stutz2020confidencecalibrated}, we use the {\em robust test error (Rerr)} metric to combine the evaluation of defenses with and without detectors.
Rerr is defined as:
\begin{equation}\label{rerr}
Rerr =
\frac{\sum^{N}_{n=1}
\max_{d(x', x) \leq \epsilon, g(x')=1} \mathbbm{1}_{f(x') \neq y}}{
\sum^{N}_{n=1} \max_{d(x', x) \leq \epsilon} \mathbbm{1}_{g(x')=1}}
\end{equation}
where $g\colon \sX \rightarrow \{0,1\}$ is a detector that accepts a sample if $g(x')=1$,
and $\mathbbm{1}_{f(x') \neq y}$ evaluates to one if $x'$ causes a misprediction and to zero otherwise.
The numerator counts the number of samples that are both accepted and lead to a successful attack (including cases where the original $x$ is incorrect), and the denominator counts the number of samples not rejected by the detector.
A defense without a detector (i.e., $g(x')=1$) reduces \Eqref{rerr} to the standard Rerr. We define {\em robust accuracy} as $1 -$ Rerr.

Note however that Rerr defined in \Eqref{rerr} has intractable maximization problem in the denominator, so \Eqref{rerr_empirical} is the empirical equation used to give an upper bound evaluation of Rerr. This empirical evaluation is the same as the evaluation in \citet{stutz2020confidencecalibrated}.

\begin{equation}
\label{rerr_empirical}
Rerr = \frac{\sum^{N}_{n=1}max\{ \mathbbm{1}_{f(\vx_{n}) \neq y_{n}} g(\vx_{n}), \mathbbm{1}_{f(\vx'_{n})\neq y_{n}} g(\vx'_{n}) \}}{\sum^{N}_{n=1}max\{ g(\vx_{n}), g(\vx'_{n}) \}} 
\end{equation}

\paragraph{Detectors} 
For a network $f$ with a detector $g$, the criterion function $c$ is misclassification with the detectors, and it is applied in line 3 in \Algref{seqattack}. This formulation enables per-example worst-case evaluation for detector defenses.

Note that we use a zero knowledge detector model, so none of the attacks in the search space are aware of the detector. However, \tool{} search adapts to the detector defense by choosing attacks with higher scores on the detector defense, which for \scode{A4}, \scode{C15} and \scode{C24} does lead to lower robustness.

\begin{figure}[h]
\begin{tikzpicture}

\node at (0.5,0) {
\begin{minipage}{0.4\textwidth}
\begin{equation*}
\begin{array}{c}
\displaystyle
\hat{f}(x') \neq \hat{f}(x) \\[0.2em]
g(x') = 1
\end{array}
\end{equation*}
\end{minipage}
};

\node at (-2.2,0) {
\begin{minipage}{0.4\textwidth}
\begin{equation*}
\displaystyle
\norm{x' - x}_{\infty} \leq \epsilon \;\;\; \text{s.t.} \\
\end{equation*}
\end{minipage}
};

\node[text width=2.8cm, align=center] at (-5.2,0) {\footnotesize
\textsc{Misclassification\\$L_{\infty}$ Attack \\ with Detector $g$}
};
\end{tikzpicture}
\vspace{-0.3em}
\end{figure}
\paragraph{Randomized Defenses} If $f$ has randomized component, $f(x_{n})$ in \Eqref{rerr_empirical} means to draw a random sample from the distribution. In the evaluation metrics, we report the mean of adversarial samples evaluated 10 times using $f$.


\begin{table*}[t]
\caption{Time limit (TL), attacks and losses result. Due to the cost of \scode{A10}, only one attack is searched and used. The Loss follows the format: \textbf{Loss - Targeted - Logit/Prob}. The abbreviations are defined in \Secref{sec_search_space}.}
\label{result_complete}
\begin{center}
{
\small
\begin{sc}
\begin{tabular}{@{}lccccccc@{}}
\toprule
  & TL(s) & Attack1 & Loss1 & Attack2 & Loss2 & Attack3 & Loss3 \\
\midrule
a1& 0.5 & APGD & Hinge-T-P & APGD & L1-D-P & APGD &CE-T-P\\
a2& 0.5 & APGD & Hinge-U-L & APGD & DLR-T-L & APGD & CE-D-P\\
a3& 0.5 & APGD & CE-T-P & APGD & DLR-U-L & APGD & L1-T-P\\
a4& 0.5 & FAB  & --F-L& APGD & LM-U-P & DeepFool& DLR-D-L\\
a5& 0.5 & APGD & Hinge-U-P & APGD &Hinge-U-P & PGD & DLR-T-P\\
a6& 0.5 & APGD & L1-D-L& APGD & DLR-U-L & APGD & Hinge-T-L\\
a7& 0.5 & APGD & DLR-T-P & APGD & DLR-U-L & APGD & Hinge-T-L\\
a8& 1 & APGD & L1-U-P & APGD & CE-U-P & APGD & CE-D-P\\
a9& 1 & APGD & DLR-U-L & APGD & Hinge-U-P & APGD & CE-U-L\\
a9'& 30 & NES & Hinge-U-P& - & -& - & -\\
b10& 3 & APGD & DLR-U-L & APGD & DLR-U-S & DeepFool & CE-D-P \\
b11& 3 & APGD & Hinge-T-P& DeepFool & L1-D-L & PGD & CE-D-P \\
b12& 3 & APGD & Hinge-T-P & DeepFool & Hinge-D-P & DeepFool & L1-D-L \\
b13& 3 & APGD & CE-D-L & APGD & DLR-F-P & DeepFool & CE-D-L \\
b14& 3 & APGD & Hinge-T-L& APGD &CE-U-P& C\&W & --U-L \\
c15& 2 & SQR & DLR-U-L & SQR& DLR-T-L & APGD & Hinge-U-P \\
c16& 3 & FAB & --F-L & APGD & L1-T-L & FAB & --F-L \\
c17& 3 & APGD & L1-D-P & APGD & CE-F-P & APGD & DLR-T-L\\
c18& 3 & SQR& Hinge-U-L & SQR & L1-U-L & SQR & CE-U-L \\
c19& 3 & APGD & L1-D-P & C\&W & Hinge-U-L & PGD & Hinge-T-L \\
c20& 3 & APGD & Hinge-U-L& APGD &DLR-T-L & FGSM & CE-U-P \\
c21& 3 & APGD &Hinge-U-L& APGD &DLR-T-L & FGSM & DLR-U-P \\
c22& 3 & PGD & DLR-U-P & FGSM & L1-U-P & FGSM & DLR-U-L \\
c23& 3 & APGD & L1-T-L & PGD & L1-U-P & PGD & L1-U-P \\
c24& 2 & PGD & L1-T-P & APGD & CE-T-P & APGD& L1-U-L \\

\bottomrule
\end{tabular}
\end{sc}
}
\end{center}
\vskip -0.1in
\end{table*}

\section{Discovered Adaptive Attacks} \label{appendix_discovered_att} \label{sec_result_adaptive_attack}
To provide more details on \Tabref{eval_full}, \Tabref{result_network_transformation} shows the network transformation result, and \Tabref{result_complete} shows the searched attacks and losses during the attack search.

\paragraph{Network Transformation Related Defenses} In the benchmark, there are $4$ defenses that are related to the network transformations. JPEG compression (\scode{JPEG}) applies image compression algorithm to filter the adversarial disturbances and to make the network non-differentiable. Reverse sigmoid (\scode{RS}) is a special layer applied on the model's logit output to obfuscate the gradient. Thermometer Encoding (\scode{TE}) is an input encoding technique to shatter the linearity of inputs. Random rotation (\scode{RR}) is in the family of randomized defense which rotates the input image by a random degree each time. \Tabref{result_network_transformation} shows where the defenses appear and what network processing strategies are applied.

\paragraph{Diversity of Attacks} From table \ref{result_complete}, the majority of attack algorithms searched are \scode{APGD}, which shows the attack is indeed a strong attack. The second or third attack can be a non-effective weak attack like \scode{FGSM} and \scode{DeepFool} in some cases, and the reason is that the noise in the untargeted \scode{CE} loss tie-breaker determines the choice of attack when none of the samples are broken by the searched attacks. In these cases, the arbitrary choice is acceptable as none of the other attacks are effective. The loss functions show variety, yet \scode{Hinge} and \scode{DLR} appear more often than \scode{CE} even we use \scode{CE} loss as the tie-breaker. This challenges the common practise of using \scode{CE} as the loss function by default to evaluate adversarial robustness.

\begin{table}[t]
\caption{List of network processing strategy used on relevant benchmarks. The format is \textbf{defense-policy}. The defenses are defined in \Secref{sec_result_adaptive_attack}. For layer removal policies, 1 means to remove the layer, 0 means not to remove the layer. For BPDA policies, I means identity, and C means using the network with two convolutions having ReLU activation in between. }
\label{result_network_transformation}
\vskip 0.15in
\begin{center}
\begin{small}
\begin{sc}
\begin{tabular}{llr}
\toprule
  & Removal Policies & BPDA Policies  \\
\midrule
a2& - & TE-C  \\
a3& JPEG-1 RS-1 & JPEG-I \\
a4& RR-0 & -  \\
a6& JPEG-1 RS-1 RR-1 & TE-C, JPEG-I \\
c20& JPEG-0 RS-0 & JPEG-I \\
c21& JPEG-1 RS-1 & JPEG-I \\
c22& RR-0 & -  \\
c23& RR-0 & -  \\
\bottomrule
\end{tabular}
\end{sc}
\end{small}
\end{center}
\vskip -0.1in
\end{table}

\section{Scalability Study} \label{appendix_scalability}

Here we provide details on scalability study in \Secref{sec_eval}.

We designed an extended search space with addition of 8 random attacks and 4 random losses to test the scalability of \tool. Random attack is to sample a point inside of the disturbance budget uniformly at random, and random loss is $\ell_{CE}$ with random sign. In our original search space for a single attack, the number of attacks is 8 and the number of losses is 4 ($8 \times 4$), so the extended search space ($16 \times 8$) has $4 \times$ the search space compared with the original space. In the other setting, we use half of the samples ($n=50$) to check \tool performance with halved search time. We evaluate block \scode{A} models except \scode{A9} model because of the high variance in result (around $ \pm 1.5$) due to the obfuscated nature of the defense.

We show the result in \Tabref{scalability_table}. We see a minor drop in performance with the extended search space or with half of the samples, and \tool still gives competitive evaluation in these scenarios. When increasing the number of trials to $96$ on the scaled dataset, the result reaches same performance.

The redundancy of $m=3$ attack is an explanation of \tool{} giving competitive performance in these scenarios. As long as one strong attack is found within the 3 attacks, the robustness evaluation is competitive.

\begin{table}[t]
\caption{Evaluating scalability of \tool{}. Original search space corresponds to the search space defined in Appendix B. Extended search space additionally contains 8 random attacks and 4 losses.  }
    \label{scalability_table}
	\begin{center}
	{\small
    \begin{tabular}{@{}lccccccc@{}}
        \toprule
         Net & \scode{AA} && \multicolumn{2}{c}{Original Search Space} & & \multicolumn{2}{c}{Extended Search Space} \\
         \cmidrule{4-5} \cmidrule{7-8}
         &&& Normal & n=50 & & k=64 & k=96 \\
        \midrule
        \scode{A1} & 44.78 && 44.69 & 44.93 && 44.80& 44.80 \\
        \scode{A2} & $\:\:$2.29 && $\:\:$1.96  & $\:\:$2.09  && $\:\:$2.14 & $\:\:$1.83 \\
        \scode{A3} & $\:\:$0.59 && $\:\:$0.11  & $\:\:$0.11  && $\:\:$0.11 & $\:\:$0.10 \\
        \scode{A4} & $\:\:$6.17 && $\:\:$3.04  & $\:\:$3.15  && $\:\:$3.47 & $\:\:$2.89 \\
        \scode{A5} & 22.30 && 12.14 & 12.53 && 11.65 & 11.85\\
        \scode{A6} & $\:\:$4.14 && $\:\:$3.94  & $\:\:$3.86  && $\:\:$4.43 & $\:\:$4.43\\
        \scode{A7} & $\:\:$2.85 && $\:\:$2.71  & $\:\:$2.78  && $\:\:$2.79 & $\:\:$2.76\\
        \scode{A8} & 19.82 && 11.11 & 11.52 && 13.02 & 11.09\\
		\scode{Avg} & 12.87 && $\:\:$9.96 & 10.12 && 10.30 & $\:\:$9.97\\
        \bottomrule
    \end{tabular}
    }
    \end{center}
\end{table}

\section{Ablation Study} \label{appendix_ablation}

Here we provide details on the ablation study in \Secref{sec_eval}. 

\subsection{Attack Algorithm \& Parameters}

In the experiment setup, the search space includes four attacks (\scode{FGSM}, \scode{PGD}, \scode{DeepFool}, \scode{C\&W}) with their generic and specific parameters shown in \Tabref{attack_loss_modifiers} and \Tabref{attack_specific_param} respectively. The loss search space is fixed to the loss in the original library implementation, and the network transformation space contains only \scode{BPDA}. \textit{Robust accuracy} (Racc) is used as the evaluation metric. The best Racc scores among \scode{FGSM}, \scode{PGD}, \scode{DeepFool}, \scode{C\&W} with library default parameters are calculated, and they are compared with the Racc from the attack found by \tool.

The result in \Tabref{attack_comparison} shows the average robustness improvement is 5.5\%, up to 17.3\%. \scode{PGD} evaluation can be much stronger after tuning by \tool, reflecting the fact that insufficient parameter tuning in \scode{PGD} was a common cause to over-estimate the robustness in literature. At closer inspection, the searched attacks have larger step sizes (typically 0.1 compared with 1/40), and higher number of attack steps (60+ compared with 40).

\begin{table}[t]
\caption{Comparison with library default parameters and the searched best attack. The implementations of \scode{FGSM}, \scode{PGD}, and \scode{DeepFool} are based on \scode{FoolBox}~\citep{rauber2017foolbox} version 3.0.0, \scode{C\&W} is based on \scode{ART}~\citep{DBLP:journals/corr/abs-1807-01069} version 1.3.0. }
    \label{attack_comparison}
	\begin{center}
	{\small
    \begin{tabular}{@{}lccccccccc@{}}
        \toprule
         & \multicolumn{2}{c}{Library Impl.} & & \multicolumn{3}{c}{\tool{}} \\
        \cmidrule{2-3} \cmidrule{5-7}
        Net & Racc & Attack  & & Racc & $\Delta$ & Attack\\
        \midrule
        \scode{A1} &	47.1& \scode{C\&W} &&	47.0& \cellcolor{green!30} -0.1 &	\scode{PGD}\\
        \scode{A2} &	13.4&	\scode{PGD} &&	$\:\:$6.7& \cellcolor{green!30} -6.8 &	\scode{PGD}\\
        \scode{A3} &	35.9&	\scode{DeepFool} &&	30.3& \cellcolor{green!30} -5.6 &	\scode{PGD}\\
        \scode{A4} &	$\:\:$6.6&	\scode{DeepFool} &&	$\:\:$6.6& $\:\:$0.0 &	\scode{DeepFool}\\
        \scode{A5} &	14.5&	\scode{PGD} &&	$\:\:$8.4& \cellcolor{green!30} -6.1 &	\scode{PGD}\\
        \scode{A6} &	35.0& \scode{PGD} &&	17.3& \cellcolor{green!30} -17.7 &	\scode{PGD}\\
        \scode{A7} &	$\:\:$6.9& \scode{C\&W} &&	$\:\:$6.6& \cellcolor{green!30} -0.3 &	\scode{C\&W}\\
        \scode{A8} &	25.4& \scode{PGD} &&	14.7& \cellcolor{green!30} -10.7 & \scode{PGD}\\
        \scode{A9} &	64.7&	\scode{FGSM}&&	62.4& \cellcolor{green!30} -2.3 &	\scode{PGD}\\
        \bottomrule
    \end{tabular}
    }
    \end{center}
\end{table}

\subsection{Loss}
Figure \ref{tpe_random_loss} shows the comparison between \scode{TPE} with loss formulation and \scode{TPE} with default loss. The search space with default loss means the space containing only \scode{L1} and \scode{CE} loss, with only untargeted loss and logit output. The result shows the loss formulation gives 3.0\% improvement over the final score.

\subsection{TPE algorithm vs Random}
\begin{figure}[h!]
\begin{center}
\centerline{\includegraphics[width=0.6\columnwidth]{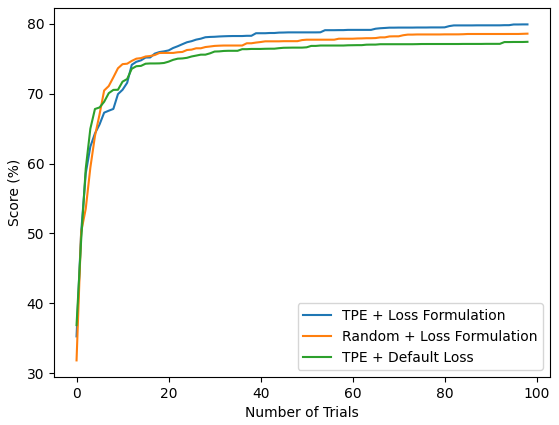}}
\vspace{-1em}
\caption{The best score progression measured by the average of 5 runs of models \scode{A1} to \scode{A9}. }
\label{tpe_random_loss}
\end{center} 
\vskip -0.3in
\end{figure}

In this experiment, we take $n=100$ samples uniformly at random and run both TPE and random search algorithm on block \scode{A} models. We record the progression of the best score in $k=100$ trials. We repeat the experiment $5$ times and average across the models and repeats to obtain the progression graph shown in \Figref{tpe_random_loss}. The result shows that \scode{TPE} finds better scores by an average of 1.3\% and up to 8.0\% (\scode{A6}).


In practice, random search algorithm is simpler and parallelizable. We observe that random search can achieve competitive performance as \scode{TPE} search.

\begin{figure}[h]
\begin{center}

\centerline{\includegraphics[trim=0 0 0 0,clip,width=0.9\columnwidth]{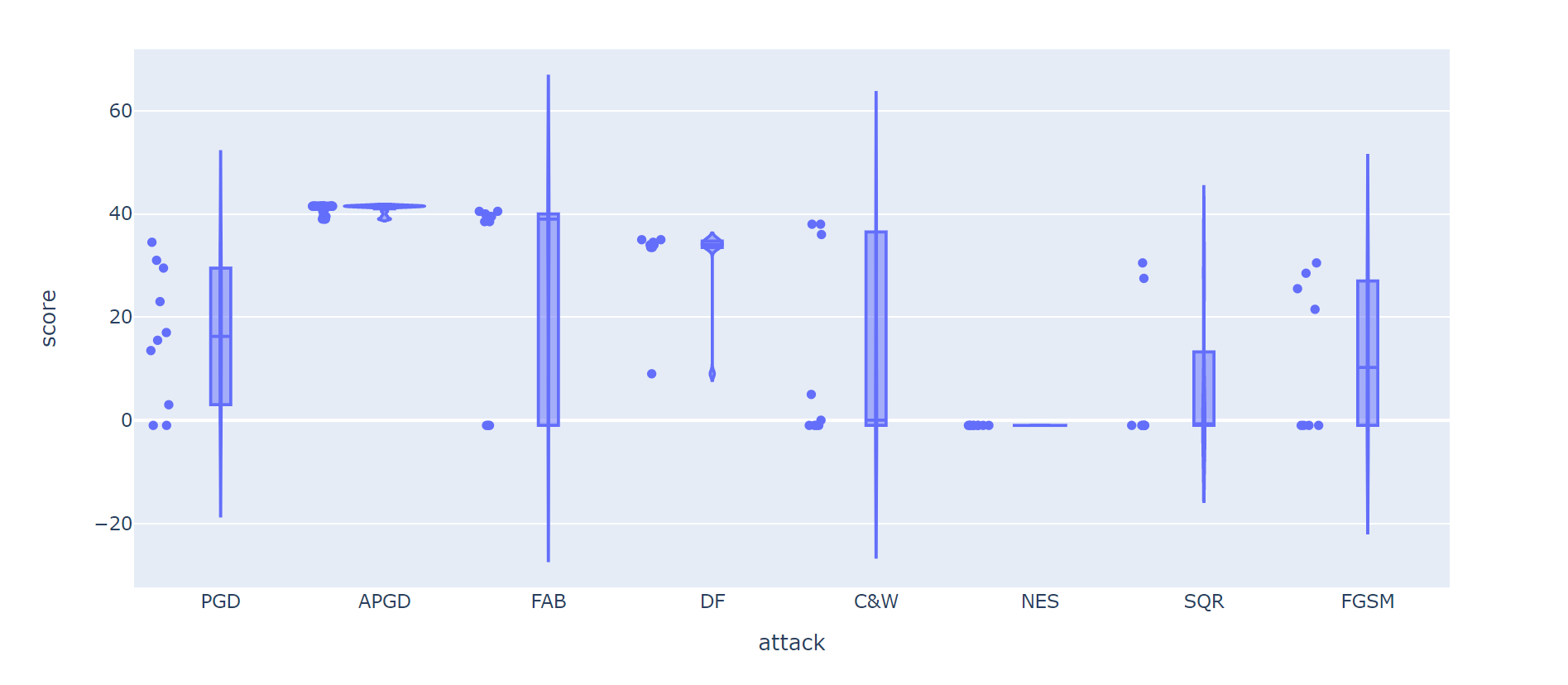}}
\vspace{-1em}
\caption{Attack-score distribution generated by \scode{TPE} algorithm on \scode{A1} model. Scores with negative values corresponds to the time-out trials.}
\label{tpe_a2}
\end{center}
\vskip -0.3in
\end{figure}

\section{Attack-Score Distribution during Search}

The analysis of attack-score distribution is useful to understand \tool. \Figref{tpe_a2} shows the distribution when running \tool on network \scode{A1}. In this experiment, the number of trials is $k=100$ and the initial dataset size is $n=200$, the time budget is $T_{c}=0.5$, and we use the search space defined in Appendix \ref{sec_search_space}. We used single GTX1060 on this experiment. We can observe the following:

\begin{itemize}

\item The expensive attack times out when $T_{c}$ values are small. Here the expensive attack \scode{NES} gets time-out because a~small $T_{c}$ is used.

\item The range and prior of attack parameters can affect the search. As we see cheap \scode{FGSM} gets time-out because the search space includes large repeat parameter.

\item Different attack algorithms have different parameter sensitivity. For examples, \scode{PGD} has a large variance of scores, but \scode{APGD} is very stable.

\item \scode{TPE} algorithm samples more attack algorithms with high scores. Here, there are 18 \scode{APGD} trials and only 7 \scode{NES} trials. \scode{TPE} favours promising attack configurations so that better attack parameters can be selected during the \scode{SHA} stage.

\item The top attacks have similar performance, which means the searched attack should have low variance in attack strength. In practice, the variance among the best searched attacks is typically small ($\pm 0.2\%$).
\end{itemize}

\section{Analysis of \tool{} Confidence Interval} \label{conf_interval}
We evaluated \tool{} using three independent runs for models in Block \scode{A} and \scode{B} as reported in \Tabref{conf_interval_table}. The result shows typically small variation across different runs (typically less than $\pm 0.2\%$), which means \tool{} is consistent for robustness evaluation.

Confidence varies across different models, and the typical reason is the variance of the attacks on the same model. For examples, models \scode{A8}, \scode{A9} are obfuscated and \scode{A5} is randomized, the attack has large variance due to the nature of these defenses.

\begin{table}[t!]
\caption{Three independent runs and confidence intervals of \tool{} for models in Block \scode{A} and \scode{B}. The bold numbers show the worst case evaluation for each model. Each confidence interval is calculated as the plus and minus the standard deviation value across the three runs. Note, that the numbers from run 3 are identical to the numbers reported in \Tabref{eval_full}.}
    \label{conf_interval_table}
	\begin{center}
	{\small
    \begin{tabular}{@{}lcccrc@{}}
        \toprule
        & \multicolumn{3}{c}{Run} & &$\sigma$\\ \cmidrule(lr){2-4} \cmidrule(lr){6-6}
        Net & 1 & 2  & 3 & & Confidence Interval \\
        \midrule
        \scode{A1} &	44.79 &	44.7 &	\textbf{44.69} & &	44.73 $\pm$	0.04 \\
		\scode{A2} &	$\:\:$2.23 &	$\:\:$2.13 &	$\:\:$\textbf{1.96} & &	$\:\:$2.11 $\pm$	0.11 \\
		\scode{A3} &	$\:\:$\textbf{0.10} &	$\:\:$0.10 &	$\:\:$0.11 & &	$\:\:$0.10 $\pm$	0.01 \\
		\scode{A4} &	$\:\:$\textbf{3.00} &	$\:\:$3.32 &	$\:\:$3.04 & &	$\:\:$3.12 $\pm$	0.14 \\
		\scode{A5} &	12.73 &	12.74 &	\textbf{12.14} & &	12.54 $\pm$	0.28 \\
		\scode{A6} &	$\:\:$4.18 &	$\:\:$4.11 &	$\:\:$\textbf{3.94} & &	$\:\:$4.08 $\pm$	0.10 \\
		\scode{A7} &	$\:\:$2.73 &	$\:\:$\textbf{2.71} &	$\:\:$2.71 & &	$\:\:$2.72 $\pm$	0.01 \\
		\scode{A8} &	10.86 &	\textbf{10.49} &	11.11 & &	10.82 $\pm$	0.25 \\
		\scode{A9} &	62.62 &	\textbf{62.31} &	63.56 & &	62.83 $\pm$	0.53 \\
		\scode{B10} &	62.80 &	62.83 &	\textbf{62.79} & &	62.81 $\pm$	0.02 \\
		\scode{B11} &	60.43 &	60.04 &	\textbf{60.01} & &	60.16 $\pm$	0.19 \\
		\scode{B12} &	\textbf{59.22} &	59.32 &	59.56 & &	59.37 $\pm$	0.12 \\
		\scode{B13} &	59.54 &	59.54 &	\textbf{59.51} & &	59.53 $\pm$	0.02 \\
		\scode{B14} &	\textbf{57.11} &	57.24 &	57.16 & &	57.17 $\pm$	0.05 \\
        \bottomrule
    \end{tabular}
    }
    \end{center}
\end{table}

\end{document}